\documentclass[10pt,twocolumn,letterpaper]{article}

\usepackage[pagenumbers]{cvpr} 

\usepackage{epsfig}
\usepackage{xcolor}
\usepackage{bm}
\usepackage{mathtools}
\usepackage{multirow}
\usepackage{makecell}
\usepackage{arydshln}

\definecolor{cvprblue}{rgb}{0.21,0.49,0.74}
\usepackage[pagebackref,breaklinks,colorlinks,citecolor=cvprblue]{hyperref}

\def\paperID{310} 
\def\confName{3DV\xspace}
\def\confYear{2024\xspace}

\DeclareRobustCommand{\colorbar}{%
  \begingroup\normalfont
  \includegraphics[height=\fontcharht\font`\B]{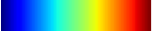}%
  \endgroup
}

\title{Diffusion Shape Prior for Wrinkle-Accurate Cloth Registration}

\author{
Jingfan Guo$^1$
\and
Fabian Prada$^2$
\and
Donglai Xiang$^3$
\and
Javier Romero$^2$
\and
Chenglei Wu$^2$
\and
Hyun Soo Park$^1$
\and
Takaaki Shiratori$^2$
\and
Shunsuke Saito$^2$
\and\\
$^1$University of Minnesota \qquad
$^2$Meta Reality Labs \qquad
$^3$Carnegie Mellon University
}

\begin{document}
\maketitle
\begin{abstract}
    Registering clothes from 4D scans with vertex-accurate correspondence is challenging, yet important for dynamic appearance modeling and physics parameter estimation from real-world data. However, previous methods either rely on texture information, which is not always reliable, or achieve only coarse-level alignment. In this work, we present a novel approach to enabling accurate surface registration of texture-less clothes with large deformation. Our key idea is to effectively leverage a shape prior learned from pre-captured clothing using diffusion models.
    We also propose a multi-stage guidance scheme based on learned functional maps, which stabilizes registration for large-scale deformation even when they vary significantly from training data.
    Using high-fidelity real captured clothes, our experiments show that the proposed approach based on diffusion models generalizes better than surface registration with VAE or PCA-based priors, outperforming both optimization-based and learning-based non-rigid registration methods for both interpolation and extrapolation tests.
\end{abstract}    

\section{Introduction}
\label{sec:introduction}

\begin{figure}[t]
    \includegraphics[width=0.9\linewidth]{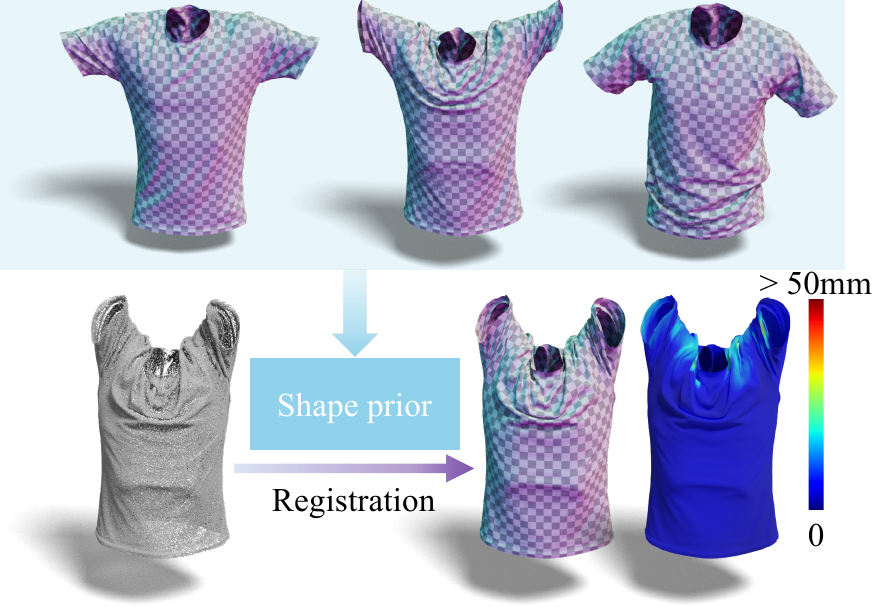}
    \centering
    \caption{Wrinkle-accurate cloth registration. We learn a strong shape prior from pre-captured 4D data using a diffusion model, and apply it to texture-less registration of the clothing with highly complex deformations.}
    \label{fig:teaser}
    \vspace{-6mm}
\end{figure}

How we dress is important in the perception of identity. The digitization of dynamically deforming clothes is, therefore, one of the core technologies to enable genuine social interaction in virtual environments. This will bring out a myriad of applications including photorealistic telepresence, virtual try-on and visual effects for game and movies. Recently, remarkable progress has been made in computer vision and graphics by modeling photorealistic appearance~\cite{xiang2022dressing} and plausible geometric deformations~\cite{patel2020tailornet}. One of the essential building blocks for these approaches is surface registration, which establishes correspondences between a template model and observed 3D reconstruction at each time frame.

Classic methods like Iterative Closest Point (ICP) registration achieve low \emph{surface} error, but suffer from in-plane sliding of the vertices due to the lack of geometric constraints. This problem makes the registration results unsuitable for learning clothes characteristics such as physical parameters or statistical models of deformations.
Current registration approaches~\cite{xiang2022dressing} avoid sliding by relying on texture, i.e., photometric consistency to establish the correspondence between the template and the observed images. Due to the reliance on the texture, the performance of the registration is highly dependent on the uniqueness and contrast of the texture. It is impossible to establish reliable correspondence for regions without salient patterns.

Hand-crafted shape priors like Laplacian~\cite{sorkine2004laplacian,nicolet2021large} and ARAP~\cite{sorkine2007rigid} can be helpful for texture-less cloth registration by regularizing the cloth deformation.
They are based on the heuristic that a shape deforms under isometry.
However, this may not apply to clothing, because cloth deformation is highly complex, including stretching and bending.
The hand-crafted shape priors also require hyperparameter tuning to achieve a balance between registration objective and regularization.
To avoid heuristics and parameter tuning, our goal is to learn a shape prior from real-world clothing deformations, which can constrain our solution space within the span of plausible clothing deformations.
Existing learning-based approaches like PCA and VAE~\cite{groueix20183d, bagautdinov2018modeling} have difficulty modeling large deformation and fine details of clothing simultaneously.
Noticing the success of diffusion probabilistic models~\cite{sohl2015deep, ho2020denoising, songdenoising} on a wide variety of challenging generative tasks~\cite{rombach2022high, chung2022improving, chung2022diffusion, poole2022dreamfusion}, we argue that the representation power of diffusion models is useful for non-rigid surface registration.

In this paper, we present a novel shape prior and illustrate how to use it for surface registration in order to achieve accurate registration of clothing under large motion even without texture information.
In particular, we employ a diffusion probabilistic model~\cite{sohl2015deep, ho2020denoising, songdenoising} to learn complex shape distributions of clothing. 
Given the target 3D point clouds, we employ approximated posterior sampling~\cite{chung2022diffusion} of the diffusion model with loss functions that optimize the surface alignment.
Unfortunately, sampling the posterior directly from scratch results in unstable coarse shape that makes further refinement meaningless.
To stably guide the surface registration process, we propose a multi-stage posterior sampling process, where the early stage of the denoising process is guided by a learning-based coarse registration approach~\cite{huang2022multiway}, and the later stage is only refined with point-to-plane errors.
In this way, the registration can avoid local minima while retaining high-fidelity wrinkles with faithful surface deformations.

Real clothing exhibits intricate deformations and interactions with human body parts, which may not be precisely synthesized by physics-based simulation.
To evaluate the accuracy of surface registration in real data, we obtain ground-truth correspondence by utilizing a state-of-the-art tracking method based on clothes with a special printed pattern~\cite{halimi2022pattern}.
Experimental results show that our method generalizes to new unseen motion of the garment it was trained on (tested on t-shirts and skirts).
In addition, our diffusion-based shape prior significantly outperforms other data-driven shape priors such as PCA and hierarchical VAE as well as state-of-the-art non-rigid registration methods.

Our method has promising applications in 3D cloth modeling.
Specifically, after learning the shape prior from a garment with a special printed pattern~\cite{halimi2022pattern}, it can be used to register garments with the same shape but different textures. This enables creating appearance models as in ~\cite{xiang2022dressing}, but without requiring the garment to be densely textured.

In summary, our contributions can be summarized as follows:
\begin{itemize}
    \item a novel diffusion-based shape prior that can effectively encode highly complex clothing geometry.
    \item a cloth registration approach that leverages the shape prior to achieve accurate cloth registration even in a texture-less setting.
    \item the first evaluation on ground truth from a wide range of motions and contact of real clothes, quantitatively exposing the accuracy of each registration method in real world scenarios.
\end{itemize}

\section{Related Work}
\label{sec:related-work}

Non-rigid 3D registration is a long-standing problem in the field of computer vision and graphics. In this section we focus on approaches relevant to cloth registration and shape priors. For a more in-depth review of non-rigid 3D registration, please refer to a survey~\cite{deng2022survey}.

\noindent\textbf{Optimization-based non-rigid tracking.}
The goal of non-rigid tracking or registration is to align a stream of unstructured input surfaces with a consistent template mesh, so that the vertices in the template encode the correspondences across different frames. Early work typically uses iterative optimization to find a template deformation which minimizes an energy function including data terms (e.g.\ target-to-template distance) and regularization terms imposing predefined constraints on the template such as smoothness~\cite{sorkine2004laplacian}.
Furukawa and Ponce~\cite{furukawa2008dense} combine rigid local patches with a non-rigid global model for markerless dense 3D tracking.
Amberg et al.~\cite{amberg2007optimal} propose local affine regularization and take optimal greedy steps for non-rigid ICP. 
Zaharescu et al.~\cite{zaharescu2009surface} propose a 3D feature detector and a 3D feature descriptor for triangle meshes that can be applied to shape matching.
Pons-Moll et al.~\cite{pons2017clothcap} and Xiang et al.~\cite{xiang2021modeling,xiang2022dressing} use heuristic objectives for non-rigid matching of a template to 4D scans of clothes.
SimulCap~\cite{yu2019simulcap} fits garment templates to a depth video stream by exerting artificial forces in a mass-spring system. Compared with these approaches that rely on hand-crafted regularization constraints, we propose a learning-based approach that effectively leverages garment-specific shape priors directly learned from high-quality ground truth, and therefore achieve more accurate registration.

\noindent\textbf{Learning-based non-rigid tracking.}
Recent advancement of deep learning has sparked interest in applying deep neural networks to the non-rigid tracking problem. Parameterizing the optimization problem with deep neural networks reduces the number of iterations (typically to a single one) since training data provide better correspondence guesses than the "closest-point" guess used in methods like ICP.

Early learning-based methods use discriminative approaches like regression forest to obtain body correspondences in depth images~\cite{taylor2012vitruvian}. 
The adoption of neural networks further avoids the need of engineering the type of features.
A similar approach~\cite{kim2021deep}, where the regression forest is replaced by soft classifiers using ResUNet~\cite{choy20194d}, achieves state-of-the-art results in the FAUST dataset~\cite{bogo2017dynamic}.
Early work on deep 3D registration like 3D-Coded~\cite{groueix20183d} was based on the PointNet~\cite{qi2017pointnet} architecture.
Numerous representations (3D voxel grids~\cite{shimada2019dispvoxnets}, basis point sets~\cite{prokudin2019efficient}, zero level sets~\cite{bhatnagar2020loopreg}, signed distance functions~\cite{bozic2021neural}) have been used to represent the inputs and outputs in these learning-based approaches.
Note that deep features can be combined with iterative solvers as in~\cite{li2020learning}, or upgraded to an end-to-end trainable system in Bozic et al.~\cite{bozic2020neural}.
Most of these methods are trained in a self-supervised manner using loss functions (e.g. point-to-plane distance) and hand-crafted priors (e.g. smoothness) similar to classical approaches, circumventing the need of accurate ground truth at the cost of fidelity to real deformations.
By comparison, our method focuses on utilizing a garment-specific shape prior directly learned from high-quality ground truth to perform more accurate registration.

\noindent\textbf{Shape prior.}
Deep learning has emerged as a powerful tool to build statistical 3D shape priors directly from data. Such data prior can be useful for various downstream tasks such as animation, reconstruction and tracking.
Functional maps~\cite{ovsjanikov2012functional,litany2017deep,huang2022multiway} are a flexible framework for isometric shape matching, where a shape can be modeled by either deterministic descriptors (e.g., Laplace-Beltrami operator) or learnable descriptors.
Different versions of mesh autoencoders (multi-scale~\cite{ranjan2018generating}, an embedded deformation layer~\cite{tretschk2020demea}, and fully convolutional~\cite{zhou2020fully}) have been used to model shape variations.
Local shape models like PatchNets~\cite{tretschk2020patchnets} and DeepLS~\cite{chabra2020deep} claim better generalizability.
Minimal Neural Atlas~\cite{low2022minimal} models a 3D shape in the parametric domain as a combination of multiple charts, enabling the learning of distortion-minimal parameterization.
While these approaches have pushed forward the accuracy of shape generation, implicit surface models lack an explicit parameterization to model in-plane sliding, and approaches like functional maps only provide coarse-level registration when applied to highly non-rigid objects such as clothing.

Learning-based shape priors are also incorporated in clothing or clothed human modeling.
Cloth has been modeled independently from the body in a few classic works~\cite{patel2020tailornet,lahner2018deepwrinkles}.
TailorNet~\cite{patel2020tailornet} jointly models the pose, shape and style of clothing, where a high frequency component is responsible for representing fine details. To increase the high frequency deteails, DeepWrinkles~\cite{lahner2018deepwrinkles} uses conditional GAN to generate high resolution normal maps. Another classic system is CAPE~\cite{ma2020learning}, which related the 3D clothes to the underlying body~\cite{loper2015smpl} through 3D displacements.
All these approaches primarily focus on the synthesis of clothing shapes under different poses, and how to use implicitly learned shape priors for surface registration remains an open question.

\noindent\textbf{Diffusion models.}
Diffusion models~\cite{sohl2015deep, ho2020denoising, songdenoising} are a class of generative models that can learn the prior from highly complex data distributions by score matching.
They have achieved state-of-the-art performance in various image-based generative tasks~\cite{rombach2022high,chung2022improving,chung2022diffusion}, including a dedicated application to clothing image manipulation~\cite{kong2023leveraging}.
Diffusion models have also been applied to 3D tasks including text-to-3D generation~\cite{poole2022dreamfusion}, human motion generation~\cite{tevet2022human}, point cloud completion~\cite{lyu2022conditional}, and stereo-based human body reconstruction~\cite{shao2022diffustereo}.
Specifically, DiffuStereo~\cite{shao2022diffustereo} is closely related to our work, which uses a conditional diffusion model to refine depth maps for high-quality human body reconstruction.
Different from DiffuStereo, which only models small residual deformations in a feed-forward fashion, we propose to estimate both large deformation and detailed deformation in a unified diffusion model by integrating it into an optimization framework.

\section{Method}
\label{sec:method}

\begin{figure*}[t]
    \vspace{-2mm}
    \includegraphics[width=0.95\linewidth]{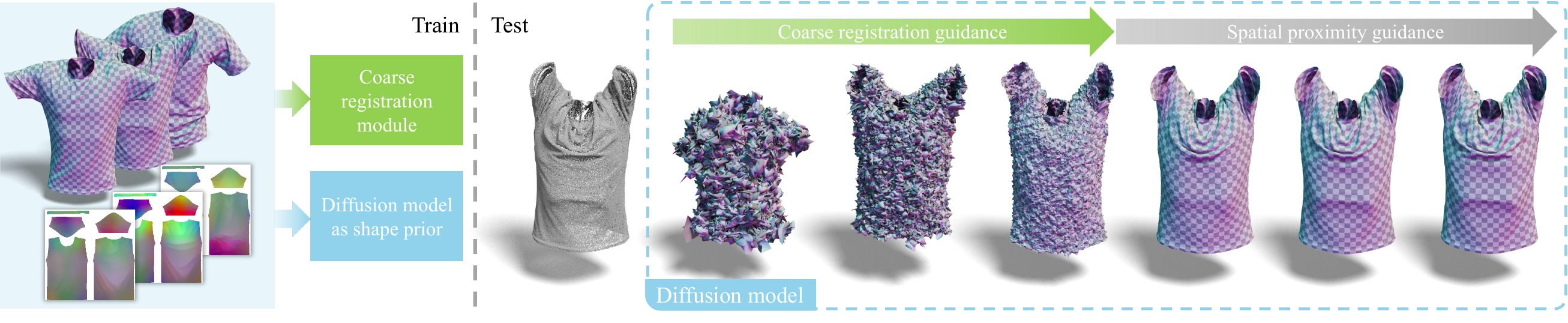}
    \centering
    \vspace{-3mm}
    \caption{We learn a diffusion-based shape prior from 4D cloth capture, and use it to accurately register the same clothing to noisy scans.}
    \vspace{-5mm}
    \label{fig:overview}
\end{figure*}

Given the ground-truth 4D scans of cloth in motion, we learn a shape prior using a diffusion model~\cite{ho2020denoising, chung2022diffusion} to simultaneously encode large deformation and fine details.
The learned shape prior can be used to register the same clothing to noisy 4D scans via multi-stage manifold guidance~\cite{chung2022improving, chung2022diffusion}.
In the early stage, our shape prior relies on coarse registration signal to achieve rough alignment.
The coarse registration signal can be acquired by markers, visual-based tracking, geometric-based tracking, or any combination of them.
In a minimum setting, where markers or visual information are not available, we rely on the geometric information by training SyNoRiM~\cite{huang2022multiway} a coarse registration module. 
In the later stage of manifold guidance, our shape prior further refines the alignment to achieve wrinkle-accurate registration by considering spatial proximity with the input 4D scan.

\subsection{Shape Representation}
We represent the clothing geometry as a 3D triangle mesh with $V$ vertices $\mathcal{V} \in \mathbb{R}^{V \times 3}$ and $F$ triangles, where the $i$-th vertex position is denoted as $\mathbf{v}_{i}$.
The $i$-th vertex corresponds with at least one point $\mathbf{u}$ on the 2D UV surface.
We define the displacement of the mesh from the mean shape with vertices $\overline{\mathcal{V}}$ as a function of UV coordinate, i.e., $\mathcal{U}|_\mathbf{u} = \mathbf{v}_i - \overline{\mathbf{v}}_i$, where $\overline{\mathbf{v}}_i$ is the $i$-th vertex of the mean shape, and $\mathcal{U}|_\mathbf{u}$ is the displacement map evaluated at $\mathbf{u}$. With an abuse of notation, we denote the mapping and its inverse as $\mathcal{U} = \Phi(\mathcal{V})$ and $\mathcal{V} = \Psi(\mathcal{U})$, respectively. Note that this parameterization is not injective, i.e., there exists a vertex that maps to multiple points in $\mathcal{U}$ where these points lie in the boundary (seam) of the unwrapped clothes.

\subsection{Diffusion-based Shape Prior}
Based on the UV displacement parameterization, we aim to learn a prior distribution of the plausible deformations, which can be used as guidance for cloth registration.
We leverage the diffusion model~\cite{ho2020denoising} that is made of two processes: forward and reverse. For the forward process, we learn a transition probability from the complete signal to a random noise $\mathbf{x}_T$ by adding noise:
\begin{align}
    \mathbf{x}_t = \sqrt{1-\beta_t} \mathbf{x}_{t-1} + \beta_t \bm\epsilon,~~~~~\mathbf{x}_0 = \mathcal{U},
\end{align}
where $\bm\epsilon \sim \mathcal{N}(\mathbf{0},\mathbf{I})$ is a sample from the Gaussian distribution. We gradually increase the variance schedule $\beta_{t}$ as increasing $t$, which reduces the impact of $\mathbf{x}_{t-1}$ while increasing that of Gaussian noise. This ensures coarse-to-fine shape learning where the large deformation (low-frequency) are modeled at large $t$ (early denoising stage), and the small deformation (high-frequency) is modeled at small $t$.

The reverse process, known as ancestral sampling~\cite{ho2020denoising}, reconstructs the signal from random noise by denoising:
\begin{align}
    \mathbf{x}_{t-1} = \frac{1}{\sqrt{\alpha_{t}}} \left( \mathbf{x}_{t} - \frac{1-\alpha_{t}}{\sqrt{1-\bar{\alpha}_{t}}} \bm{\epsilon}_{\theta}(\mathbf{x}_{t}, t) \right) + \sigma_{t} \mathbf{z} \label{eq:ancestral}
\end{align}
Based on the variance schedule $\beta_{t}$, we define $\alpha_{t} = 1 - \beta_{t}$, $\bar{\alpha}_{t} = \prod_{i=1}^{t} \alpha_{i}$, and $\sigma_{t} = \sqrt{\tilde{\beta}_{t}} = \sqrt{\frac{1-\bar{\alpha}_{t-1}}{1-\bar{\alpha}_{t}} \beta_{t}} $.
The learnable neural network $\bm{\epsilon}_{\theta}$ parameterized by $\theta$ aims to predict the noise $\bm{\epsilon}$ from corrupted data $\mathbf{x}_{t}$.
We train $\bm{\epsilon}_{\theta}$ with a weighted variational bound~\cite{ho2020denoising} as the objective:
\begin{align}
    L = \mathbb{E}_{t, \mathbf{x}_0, \bm{\epsilon}} \left[ \Vert \bm{\epsilon} - \bm{\epsilon}_{\theta} (\sqrt{\bar{\alpha}_{t}} \mathbf{x}_0 + \sqrt{1-\bar{\alpha}_{t}} \bm{\epsilon}, t) \Vert^{2} \right] 
\end{align}

\begin{figure*}[t]
    \vspace{-2mm}
    \includegraphics[width=0.9\linewidth]{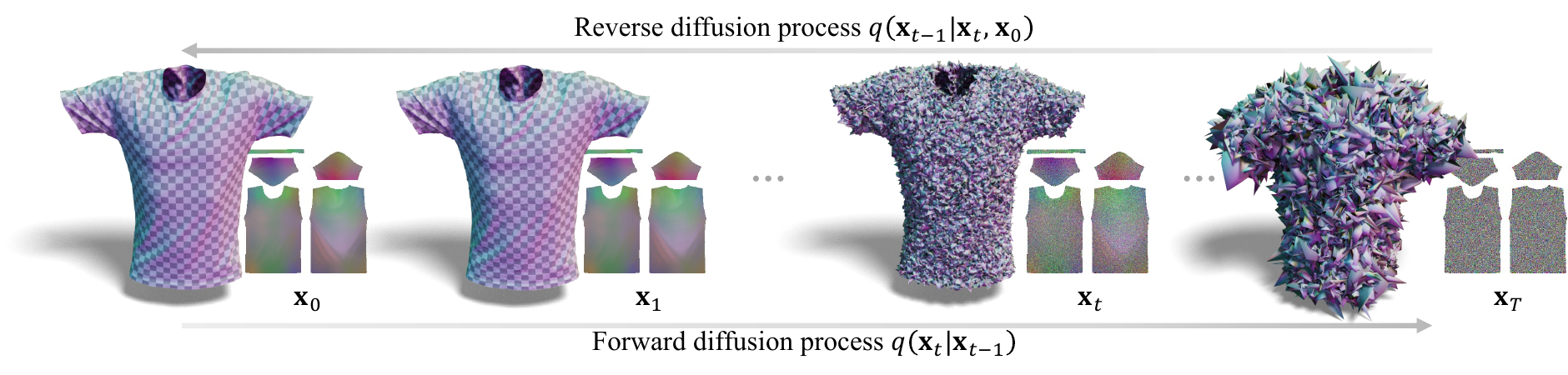}
    \centering
    \vspace{-3mm}
    \caption{Illustration of the diffusion model for clothing shape. In the forward process, we gradually add noise to the UV displacement map $\mathbf{x}_{0}$ to acquire an isotropic Gaussian distribution $\mathbf{x}_{T}$. To sample from the learned data distribution, we recover $\mathbf{x}_{0}$ by gradually denoising the corrupted UV displacement map.}
    \vspace{-5mm}
    \label{fig:diffusion}
\end{figure*}

Iterating Equation~(\ref{eq:ancestral}) will generate a plausible data sample $\mathbf{x}_{0}$ from the learned prior.
Figure~\ref{fig:diffusion} illustrates the forward and reverse diffusion processes.

\begin{figure}[t]
    \includegraphics[width=0.9\linewidth]{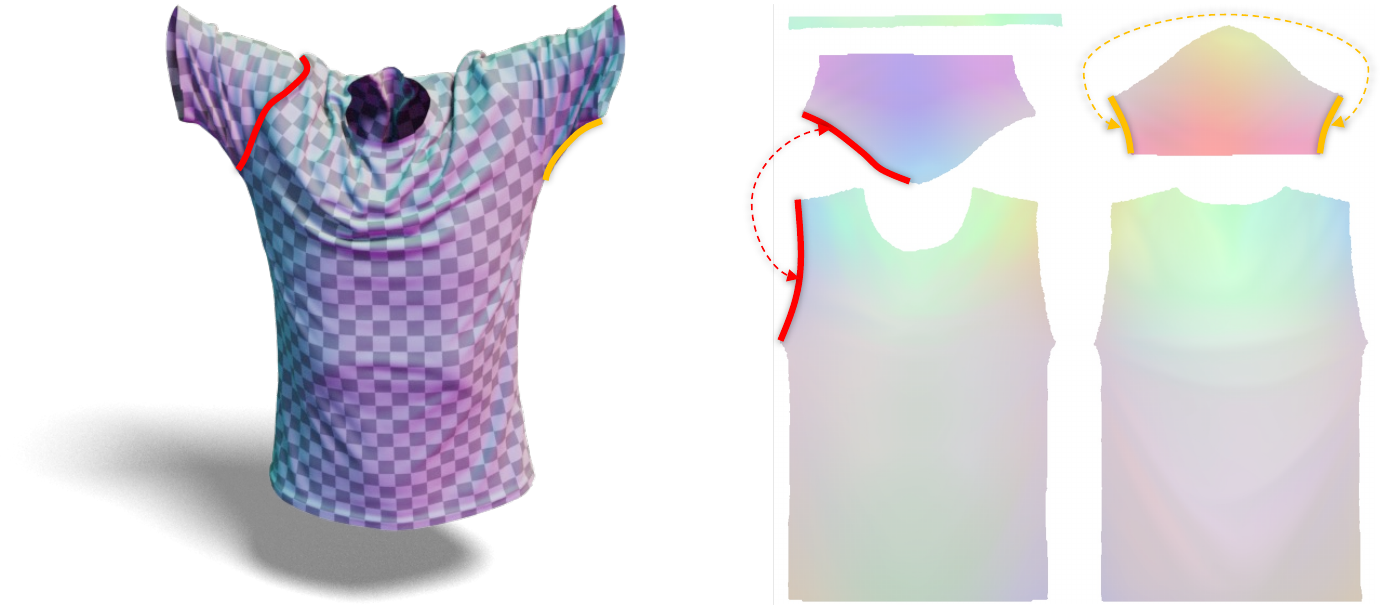}
    \centering
    \caption{Seam stitching. We enforce the same noise value for corresponding points on the seams in the reverse process.}
    \vspace{-6mm}
    \label{fig:seam_demo}
\end{figure}

\noindent\textbf{Seam stitching.}
As illustrated in Figure~\ref{fig:seam_demo}, panels of the clothing can be disconnected along seams in UV parameterization.
A plausible clothing shape has a smooth surface that maps to smoothly transitioned values across seams.
To avoid the clothing being separated apart at the seam, at every time step in the reverse process, we enforce the noise value on the seams to be the same for corresponding points by $\mathbf{x}_{t-1}^{'} = \Phi(\Psi(\mathbf{x}_{t-1}))$.
Note that the mapping $\Psi$ from UV to mesh space is not injective, so the ambiguity is solved by averaging the 3D locations of UV positions which refer to the same point. 

\subsection{Non-rigid Registration via Manifold Guidance}
The ancestral sampling in Equation~(\ref{eq:ancestral}) allows generating diverse plausible shape deformations. 
To ensure that the deformed shape matches the visible surface of the clothing, it is necessary to steer the reverse diffusion process.
Noticing that the gradient of log marginal density can be approximated by the learned network $\nabla_{\mathbf{x}_{t}} \log p(\mathbf{x}_{t}) \simeq - \bm{\epsilon}_{\theta} / \sigma_{t}$, we make use of manifold guidance~\cite{chung2022improving,chung2022diffusion} to sample the near optimal shape deformation that can match the observed points. The manifold guidance maximizes the log-likelihood of every diffusion state, $\mathbf{x}_t$, given the observation $\mathcal{Y}\in\mathbb{R}^{P\times 3}$ where $P$ is the number of observed 3D points:
\begin{equation}
    \nabla_{\mathbf{x}_{t}} \log p(\mathbf{x}_{t} | \mathcal{Y}) \simeq - \frac{\bm{\epsilon}_{\theta}(\mathbf{x}_{t}, t)}{\sigma_{t}} - \rho \nabla_{\mathbf{x}_{t}} d (\hat{\mathbf{x}}_{0}, \mathcal{Y})
\end{equation}
where $d$ is a distance measurement in Euclidean space, and $\rho$ is the step size of guidance.
The posterior mean $\hat{\mathbf{x}}_{0}$ can be estimated from $\mathbf{x}_{t}$
\begin{equation}
    \hat{\mathbf{x}}_{0} = \frac{1}{\sqrt{\bar{\alpha}_{t}}} \mathbf{x}_{t} - \sqrt{\frac{1-\bar{\alpha}_{t}}{\bar{\alpha}_{t}}} \bm{\epsilon}_{\theta}(\mathbf{x}_{t}, t)
    \label{eq:x0_hat}
\end{equation}

We take a multi-stage distance measurement $d$ with the decreasing of time step $t$ in the reverse diffusion process
\begin{equation}
    d (\hat{\mathbf{x}}_{0}, \mathcal{Y}) = \left\{
        \begin{array}{ll}
            \sum_{i=1}^{V} \varrho \left( \left\Vert \tilde{\mathbf{v}}_{i} - \hat{\mathbf{v}}_{i} \right\Vert \right) & t > \tau \\
            \sum_{i=1}^{P} \varrho \left( \left( \mathbf{y}_{i} - \mathfrak{N} (\hat{\mathcal{V}}, \mathbf{y}_{i}) \right)^{\intercal} \mathbf{n}_{\mathbf{y}_{i}} \right)  & t \le \tau
        \end{array}\right.
    \label{eq:distance}
\end{equation}
where $\tilde{\mathbf{v}}_{i}$ is the target position predicted by the coarse registration module $\mathcal{C}(\mathcal{Y})$, and $\varrho (\cdot)$ is the Huber robust function~\cite{huber1992robust}.
The mesh vertices are mapped from the UV displacement map as $\hat{\mathcal{V}} = \Phi(\hat{\mathbf{x}}_{0})$ with the $i$-th vertex denoted as $\hat{\mathbf{v}}_{i}$.
The time step $\tau$ is the point where the distance measurement is changed.
$\mathfrak{N}$ retrieves the closest point in $\hat{\mathcal{V}}$, and $\mathbf{n}_{\mathbf{y}_{i}} \in \mathbb{R}^{3}$ is the normal vector of $\mathbf{y}_{i}$.

When $t > \tau$, we use the signals from the coarse registration module $\mathcal{C}$ to guide the large deformation of the clothing shape, until we achieve a rough alignment with the input point cloud.
We adopt SyNoRiM~\cite{huang2022multiway} as the coarse registration module $\mathcal{C}$.
It learns to predict per-vertex 3D flow from template to target shape given a 3D point cloud as input.
The dense correspondences it infers are not strictly necessary since our method works well with sparse ones as shown in Section~\ref{sec:experiments}.
After $ t \le \tau$, the vertices are guided by point-to-plane errors~\cite{chen1992object} based on spatial proximity from the point $\mathbf{y}_{i}$, which is typically used for non-rigid registration tasks.
The point-to-plane distance helps to avoid overestimating the distance when the input point cloud contains holes.

After reaching $t=0$ in the reverse diffusion process, we repeat the final denoising step with point-to-plain guidance to adjust the inferred vertices to the high-frequency surface details of the point cloud.

\section{Experiments}
\label{sec:experiments}

We evaluate our method by comparing with baseline methods quantitatively and qualitatively, demonstrating the effectiveness of the proposed approach.
We further discuss the design choice of our method in the ablation study and showcase that our method can be flexible in practical use cases when sparse tracking signals are available.
See more experiments in Supp.\ Mat., including robustness to noisy input and cross-subject generalization.

\subsection{Dataset and Settings}
\noindent\textbf{Dataset.}
We conduct our experiments using the pattern-based cloth registration dataset~\cite{halimi2022pattern}, which provides a template geometry for each clothing type, as well as accurate registrations in the same topology.
We use the provided data as ground-truth for both training and evaluation.
As the dataset does not release the original point clouds, we instead construct partial 3D reconstruction to be used as input at test time.
See Supp.\ Mat.\ for details.

Since the body is not included in the dataset, we use the ground-truth registration to compute the global translation and rotation of each frame w.r.t.\ the mean shape using Procrustes analysis, and then normalize the data by applying inverse of the global transformation on each frame.
In real use cases where the body is available, we can apply a similar normalization by estimating the body pose, and taking the pelvis transformation as global transformation.

We use T-shirt on "subject\_00" (\emph{T-shirt 1}), T-shirt on "subject\_04" (\emph{T-shirt 2}, with a stiffer material than \emph{T-shirt 1}), skirt on "subject\_03" (\emph{Skirt 1}), and skirt on "subject\_04" (\emph{Skirt 2}, longer than \emph{Skirt 1}) in our experiments. 
For each data sequence, we split the frames into training set and test set, which further includes interpolation and extrapolation sets.
The interpolation test set is uniformly sampled from the entire sequence, so its data distribution is similar to the training set.
The extrapolation test set is a manually selected short sequence consisting of body poses unseen in training set.
All learning-based methods use this identical train-test split.

\noindent\textbf{Implementation details.}
We train the diffusion model with $T=1,000$ steps, and sample a subset $S=50$ steps using DDIM~\cite{songdenoising} with $\eta=0$ in the reverse process.
We use a linear variance schedule that increases from $\beta_1 = 10^{-4}$ to $\beta_T = 0.02$.
The network $\bm{\epsilon}_{\theta}$ is implemented as a U-Net~\cite{ronneberger2015u} that takes in a $256 \times 256$ UV displacement map.
The invalid pixels on the UV map are masked out during both training and testing.
We use SyNoRiM~\cite{huang2022multiway} as the coarse registration module in our pipeline.
It is trained in a pairwise manner between the mean template shape and each training sample with randomly sub-sampled mesh vertices.
Note that our method can seamlessly integrate with not only SyNoRiM but also any coarse registration methods.

\noindent\textbf{Metrics.}
We quantitatively evaluate the performance of our method and baseline methods using vertex error $E_{v}$ and bidirectional point-to-plane error $E_{pt}$ and $E_{ps}$.
\begin{align}
    E_{v} &= \frac{1}{V} \sum_{i=1}^{V} \left\Vert \hat{\mathbf{v}}_i - \mathbf{v}_i \right\Vert \\
    E_{pt} &= \frac{1}{V} \sum_{i=1}^{V} \left\vert \left( \mathbf{v}_{i} - \mathfrak{N} (\hat{\mathcal{V}}, \mathbf{v}_{i}) \right)^{\intercal} \mathbf{n}_{\mathbf{v}_{i}} \right\vert \label{eq:point-to-plane} \\
    E_{ps} &= \frac{1}{V} \sum_{i=1}^{V} \left\vert \left( \hat{\mathbf{v}}_{i} - \mathfrak{N} (\mathcal{V}, \hat{\mathbf{v}}_{i}) \right)^{\intercal} \mathbf{n}_{\hat{\mathbf{v}}_{i}} \right\vert
\end{align}
Our actual goal is to achieve a low $E_{v}$ for accurate alignment, while $E_{pt}$/$E_{ps}$ are also important. $E_{v}$ directly indicates the accuracy of the registration, while $E_{pt}$/$E_{ps}$ measures only \emph{surface} alignment.
When $E_{pt}$/$E_{ps}$ is small, $E_{v}$ can still be large due to in-plane sliding. Similarly, lower $E_{v}$ with higher $E_{pt}$/$E_{ps}$ is not preferable due to large deviation from the true surface.
Our goal is to achieve low $E_{v}$ with reasonable $E_{pt}$/$E_{ps}$.

\begin{table*}[ht]
    \setlength{\tabcolsep}{1.2pt}
    \centering
    \footnotesize{
    \begin{tabular}{l | c  c ;{2pt/2pt} c  c | c  c ;{2pt/2pt} c  c | c  c ;{2pt/2pt} c  c | c  c ;{2pt/2pt} c  c } 
        \hline
        & \multicolumn{4}{c|}{\emph{T-shirt 1}} & \multicolumn{4}{c|}{\emph{T-shirt 2}} & \multicolumn{4}{c|}{\emph{Skirt 1}} & \multicolumn{4}{c}{\emph{Skirt 2}} \\
        & \multicolumn{2}{c;{2pt/2pt}}{Int. set} & \multicolumn{2}{c|}{Ext. set} & \multicolumn{2}{c;{2pt/2pt}}{Int. set} & \multicolumn{2}{c|}{Ext. set} & \multicolumn{2}{c;{2pt/2pt}}{Int. set} & \multicolumn{2}{c|}{Ext. set} & \multicolumn{2}{c;{2pt/2pt}}{Int. set} & \multicolumn{2}{c}{Ext. set} \\
        \cline{2-17}
        & $E_{v}$ & $E_{pt}$/$E_{ps}$  & $E_{v}$  & $E_{pt}$/$E_{ps}$  & $E_{v}$  & $E_{pt}$/$E_{ps}$  & $E_{v}$  & $E_{pt}$/$E_{ps}$ & $E_{v}$ & $E_{pt}$/$E_{ps}$  & $E_{v}$  & $E_{pt}$/$E_{ps}$ & $E_{v}$ & $E_{pt}$/$E_{ps}$  & $E_{v}$  & $E_{pt}$/$E_{ps}$  \\
        \hline
        SyNoRiM~\cite{huang2022multiway} & 5.76 & 1.37/1.68 & 11.13 & 1.38/1.67 & 6.32 & 1.39/1.80 & 10.09 & 1.41/1.80 & 18.94 & 1.57/2.76 & 24.05 & 1.61/2.87 & 20.88 & 1.68/2.70 & 22.17 & 1.71/2.67  \\
        Lap. reg.~\cite{sorkine2004laplacian} & 5.04 & 0.65/0.64 & 10.80 & 0.66/0.64 & 5.60 & 0.59/0.62 & 9.61 & 0.59/0.62 & 18.33 & 0.71/0.73 & 23.71 & 0.73/0.73 & 20.47 & 0.73/0.72 & 21.82 & 0.66/0.67 \\
        Lap. precond.~\cite{nicolet2021large} & 5.00 & 0.53/0.59 & 10.64 & 0.53/0.59 & 5.53 & 0.52/0.59 & 9.44 & 0.51/0.58 & 18.09 & 0.55/0.66 & 23.40 & 0.56/0.67 & 20.42 & 0.60/0.70 & 21.84 & 0.57/0.67 \\
        ARAP reg.~\cite{sorkine2007rigid} & 4.57 & 0.52/0.60 & 10.48 & 0.51/0.60 & 5.16 & 0.53/0.60 & 9.23 & 0.51/0.59 & 17.88 & 0.59/0.75 & 23.36 & 0.60/0.76  & 19.88 & 0.65/0.77 & 21.25 & 0.58/0.68 \\
        \hline
        3D-CODED~\cite{groueix20183d} & 11.02 & 2.50/3.31 & 21.19 & 3.01/5.47 & 13.24 & 2.93/5.15 & 14.76 & 3.45/5.22 & 18.44 & 2.03/5.21 & 24.30 & 2.25/6.33 & 17.78 & 3.37/6.09 & 28.25 & 4.51/9.06 \\        
        3D-CODED opt. & 10.26 & 2.41/3.06 & 18.81 & 2.72/4.48 & 11.25 & 2.67/4.09 & 13.72 & 2.95/4.07 & 18.14 & 1.98/4.83 & 23.47 & 2.18/5.61  & 17.07 & 3.05/5.09 & 25.40 & 3.92/6.72 \\
        \hline
        PCA & 10.24 & 2.62/2.93 & 14.30 & 2.95/3.75 & 8.69 & 2.99/3.41 & 12.96 & 3.58/4.36 & 15.87 & 3.36/4.10 & 21.12 & 4.10/5.09  & 17.79 & 4.87/5.27 & 24.44 & 6.22/7.09 \\
        \hline
        Comp. VAE~\cite{bagautdinov2018modeling} & 5.05 & 0.72/0.78 & 10.74 & 0.73/0.79 & 5.67 & 0.85/0.93 & 9.63 & 0.88/0.95 & 18.04 & 0.64/0.72 & 23.47 & 0.65/0.74  & 20.57 & 0.79/0.83 & 21.94 & 0.79/0.86 \\
        \hline
        Ours & \textbf{3.16} & 0.57/0.62 &  \textbf{9.51} & 0.61/0.75  & \textbf{3.94} & 0.57/0.63 & \textbf{8.59} & 0.61/0.76 & \textbf{15.07} & 0.65/0.73  & \textbf{21.01} & 0.73/0.78  & \textbf{16.66} & 0.71/0.78 & \textbf{19.71} & 0.67/0.75 \\
        \hline
    \end{tabular}
    }
    \vspace{-2mm}
    \caption{Quantitative comparison to baseline methods. Error metrics are measured in mm. \textbf{Bold} indicates the best $E_{v}$. Our goal is to achieve low vertex error $E_{v}$ with reasonable point-to-plane error $E_{pt}$ and $E_{ps}$.}
    \label{tab:baselines}
    \vspace{-5mm}
\end{table*}

\subsection{Comparison to Baseline Methods}
\begin{figure*}[t]
    \includegraphics[width=0.96\linewidth]{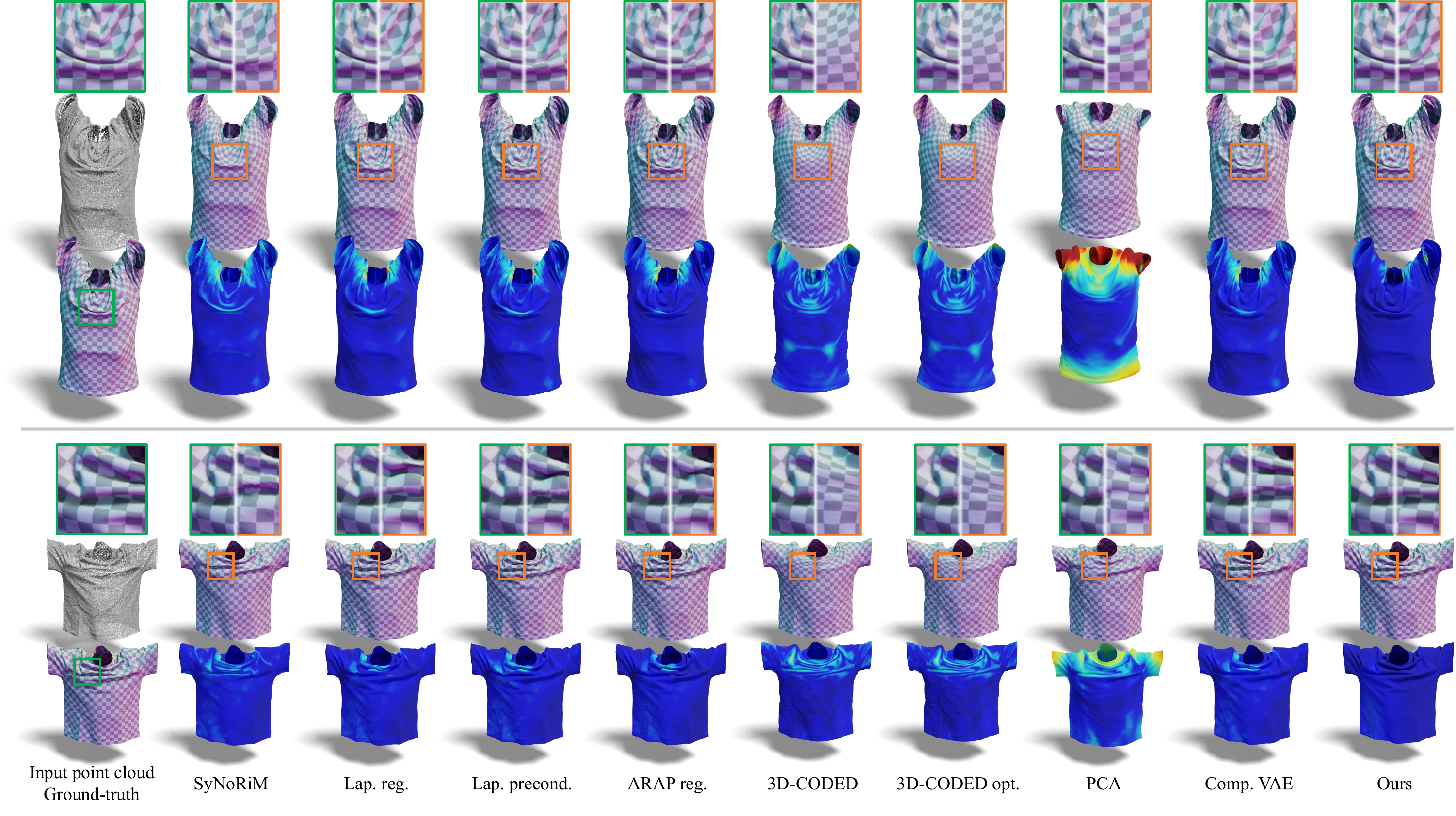}
    \centering
    \caption{Comparison to baseline methods on \emph{T-shirt 1} and \emph{T-shirt 2}. In each example, the middle-left is the input point cloud, the bottom-left is the ground-truth, the top-left is zoom-in view of ground-truth. The rest are the results of different methods, where the top row shows side-by-side comparison to ground-truth, the middle row shows the geometry with normal rendering, while the bottom row shows vertex error $E_{v}$ in color ($0mm$ \colorbar{} $>50mm$).}
    \label{fig:comparison_tshirts}
    \vspace{-3mm}
\end{figure*}

\begin{figure*}[t]
    \includegraphics[width=0.96\linewidth]{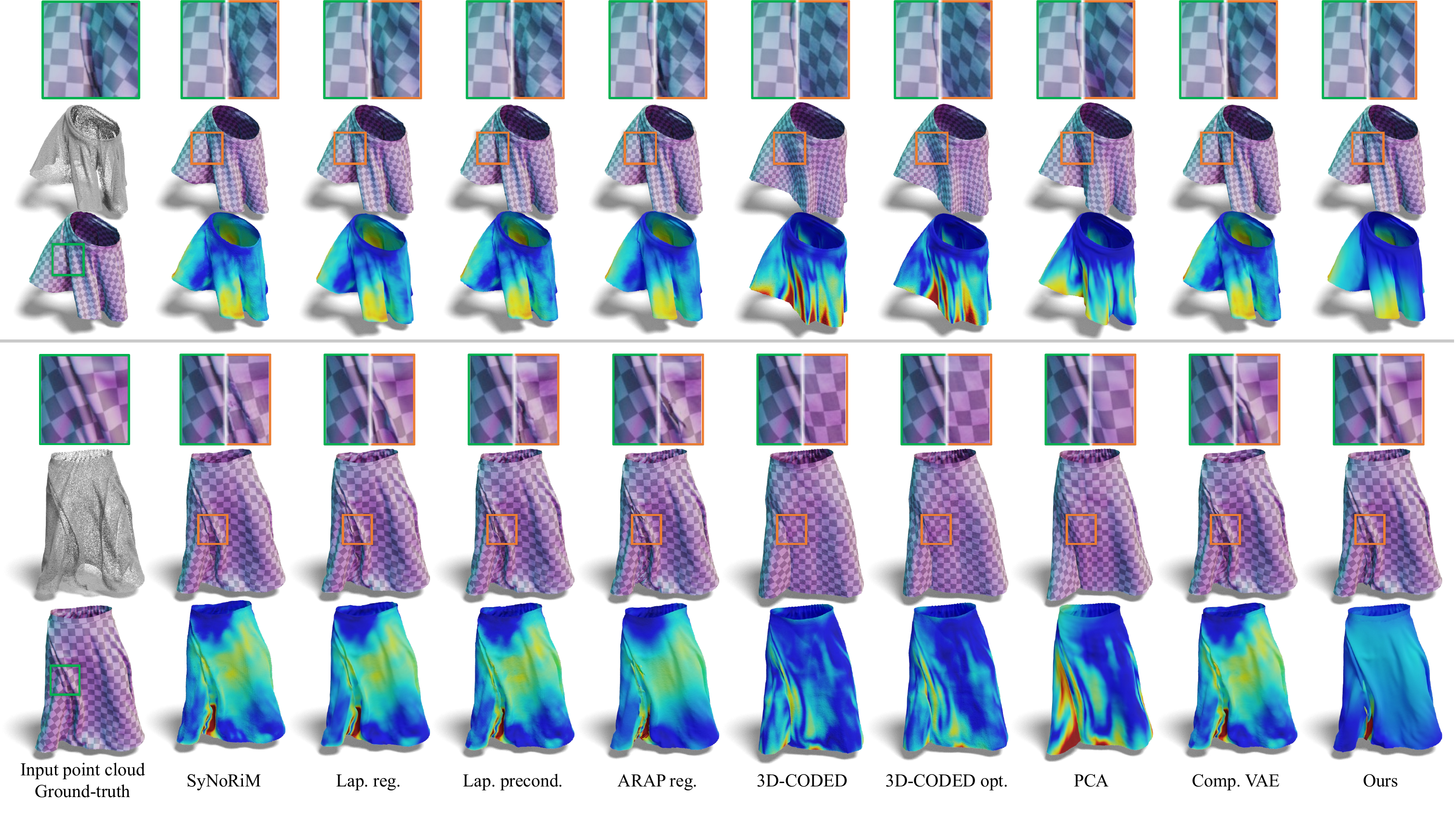}
    \centering
    \caption{Comparison to baseline methods on \emph{Skirt 1} and \emph{Skirt 2}. See Figure~\ref{fig:comparison_tshirts} for explanation, and Supp.\ Mat.\ for more results.}
    \label{fig:comparison_skirts}
    \vspace{-3mm}
\end{figure*}

SyNoRiM~\cite{huang2022multiway} is a general-purpose non-rigid registration method.
Since SyNoRiM tends to produce over-smoothed results lacking fine details like wrinkles, we further refine SyNoRiM predictions by optimizing point-to-plane distance together with heuristic shape priors (Laplacian~\cite{sorkine2004laplacian,nicolet2021large} and ARAP~\cite{sorkine2007rigid}) as in classical ICP methods.
In Table~\ref{tab:baselines}, we quantitatively show that our full pipeline consistently outperforms SyNoRiM and its heuristic refinement on the metric of vertex error, which is our main goal.
Qualitative results in Figure~\ref{fig:comparison_tshirts}, \ref{fig:comparison_skirts} show that our method consistently produces better registration with lower vertex error and realistic wrinkles.

\begin{figure}[t]
    \includegraphics[width=0.97\linewidth]{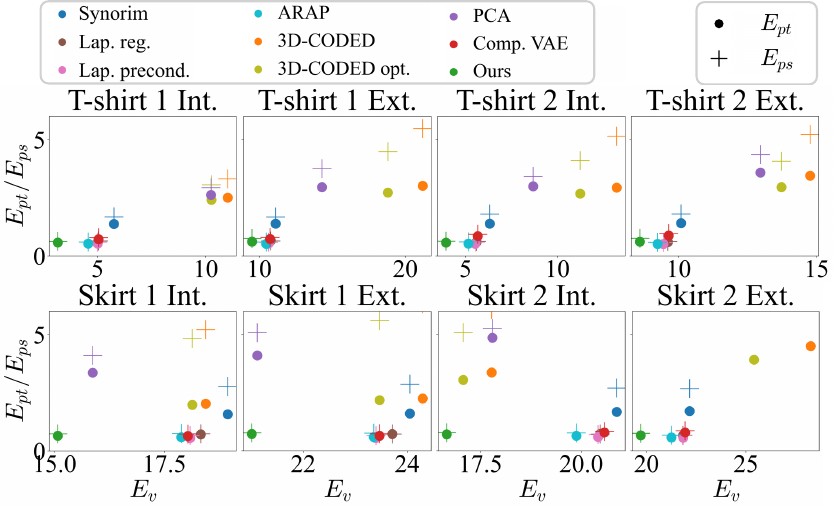}
    \centering
    \vspace{-3mm}
    \caption{Error plot of $E_{v}$ vs.\ $E_{pt}$/$E_{ps}$. Closer to origin (lower $E_{v}$ and $E_{pt}$/$E_{ps}$) is a preferred solution.}
    \label{fig:error_plot}
    \vspace{-3mm}
\end{figure}

We also compare our approach to data-driven shape priors 3D-CODED~\cite{groueix20183d}, PCA, and compositional VAE~\cite{bagautdinov2018modeling}.
In 3D-CODED, we model the 3D point translation and 3D patch deformation following the original setting.
The 3D-CODED results can be further refined by optimizing Chamfer distance (denoted 3D-CODED opt.).
For PCA, we model the per-vertex 3D displacement from mean shape, and keep a number of principal components that retains $95\%$ explained variance.
At test time, we estimate the PCA coefficients by solving least squares to dense target points given by SyNoRiM.
For compositional VAE, we model the UV displacement map similar to our setting.
At test time, we initialize the latent code by feeding the SyNoRiM result to the encoder, then optimize the latent code by minimizing point-to-plane distance in Equation~(\ref{eq:point-to-plane}).
Table~\ref{tab:baselines} shows that our shape prior is more effective than baseline data-driven shape priors.
Please note that PCA achieves comparable $E_{v}$ to our method on \emph{Skirt 1} and \emph{Skirt 2}, but it shows significantly worse plane error $E_{pt}$ and $E_{ps}$.
As a linear model, PCA may not be suitable for this inherently nonlinear problem, so it is not flexible enough to achieve accurate surface-level alignment.
It cannot fit to large deformations that are far from the mean shape as shown in Figure~\ref{fig:comparison_tshirts}, \ref{fig:comparison_skirts}, even though they are from the interpolation set and close to some training samples.
Comparing to all baseline methods, Figure~\ref{fig:error_plot} shows that our method consistently achieves the best balance of lower $E_{v}$ and $E_{pt}$/$E_{ps}$.

\begin{table}[t]
    \centering
    \small{
    \begin{tabular}{l | c  c | c  c } 
        \hline
        & \multicolumn{4}{c}{\emph{T-shirt 1}} \\
        & \multicolumn{2}{c|}{Interpolation set} & \multicolumn{2}{c}{Extrapolation set} \\
        \cline{2-5}
        & $E_{v}$ & $E_{pt}$ / $E_{ps}$  & $E_{v}$  & $E_{pt}$ / $E_{ps}$  \\
        \hline
        $\tau=40$ & 32.16 & 0.67 / 1.29 & 36.46 & 0.72 / 1.93 \\
        $\tau=30$ & 3.39 & 0.57 / 0.62 & 9.40 & 0.61 / 0.73 \\
        $\tau=20$ & 3.16 & 0.57 / 0.62 & 9.51 & 0.61 / 0.75 \\
        $\tau=10$ & 3.47 & 0.58 / 0.64 & 9.82 & 0.63 / 0.79 \\
        $\tau=0$ & 3.82 & 0.59 / 0.67 & 10.10 & 0.66 / 0.86 \\
        \hline
    \end{tabular}
    }
    \vspace{-2mm}
    \caption{The effect of the guidance breakpoint $\tau$. A large $\tau$ significant impair the performance.}
    \label{tab:tau}
    \vspace{-3mm}
\end{table}

\subsection{Ablation Study}
\noindent\textbf{Guidance breakpoint $\tau$.}
In Table~\ref{tab:tau}, we quantitatively show the impact of varying the guidance breakpoint $\tau$ in Equation~(\ref{eq:distance}) on the \emph{T-shirt 1} sequence.
The performance with large $\tau$ is significantly worse, indicating that the first stage of the manifold guidance plays a key role, and a coarse registration module is necessary in our method.
When $\tau$ is in a reasonable range, it does not significantly affect the performance, although a too small $\tau$ amplifies the influence of error from the coarse registration.

\begin{figure}[t]
    \includegraphics[width=\linewidth]{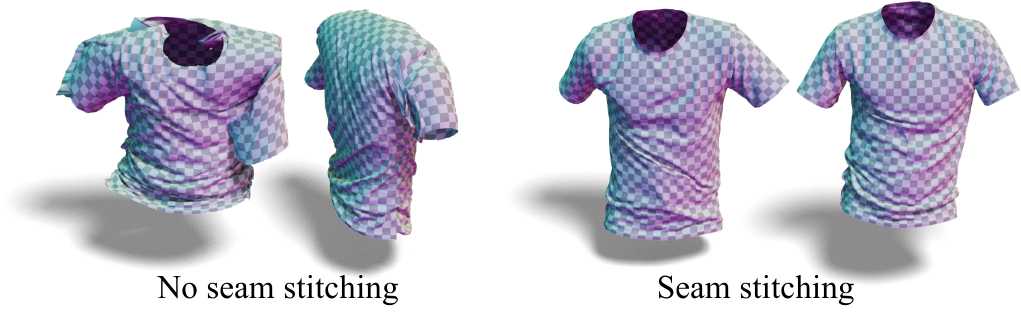}
    \centering
    \vspace{-5mm}
    \caption{The effect of seam stitching strategy. Without seam stitching, the generated clothing shape may have separated parts, as continuity is not enforced across seams.}
    \label{fig:seam}
    \vspace{-5mm}
\end{figure}

\noindent\textbf{Seam stitching.}
Given a 2D parameterization with multiple islands for the clothing, it is important to enforce the continuity across the seams.
Figure~\ref{fig:seam} illustrates that the proposed seam stitching strategy prevents generating implausible shape with separated clothing parts.

\begin{table}[t]
    \centering
    \small{
    \begin{tabular}{l | c | c  c | c  c } 
        \hline
        & \multirow{3}{*}{$N_{k}$} & \multicolumn{4}{c}{\emph{T-shirt 1}} \\
        & & \multicolumn{2}{c|}{Interpolation set} & \multicolumn{2}{c}{Extrapolation set} \\
        \cline{3-6}
        & & $E_{v}$ & $E_{pt}$ / $E_{ps}$  & $E_{v}$  & $E_{pt}$ / $E_{ps}$  \\
        \hline
        \multirow{3}{*}{PCA} & 50 & 10.65 & 2.67 / 3.08 & 13.86 & 2.98 / 3.87 \\
        & 100 & 10.22 & 2.60 / 2.94 & 13.36 & 2.91 / 3.72 \\
        & 200 & 10.28 & 2.61 / 2.97 & 13.55 & 2.95 / 3.78 \\
        \hline
        \multirow{3}{*}{\makecell{Comp. \\ VAE}} & 50 & 30.08 & 1.67 / 6.00 & 37.16 & 1.45 / 8.23 \\
        & 100 & 28.41 & 1.24 / 5.94 & 34.55 & 1.07 / 7.93 \\
        & 200 & 25.41 & 1.03 / 5.67 & 30.54 & 0.87 / 7.44 \\
        \hline
        \multirow{3}{*}{Ours} & 50 & 3.59 & 0.58 / 0.65 & 6.10 & 0.62 / 0.79 \\
        & 100 & 2.62 & 0.57 / 0.62 & 4.89 & 0.61 / 0.74 \\
        & 200 & 2.39 & 0.57 / 0.62 & 4.53 & 0.61 / 0.74 \\
        \hline
    \end{tabular}
    }
    \vspace{-3mm}
    \caption{Sparse ground-truth guidance. Our method works with sparse tracking signals. PCA performs similarly to the original setting, while the performance of compositional VAE significantly decreases.}
    \label{tab:application}
    \vspace{-3mm}
\end{table}

\subsection{Application}
In real-world scenarios, clothing usually comes with textures that make visual keypoint tracking possible.
Our method is flexible in that it can take advantage of such information when available.
To mimic the use case where keypoint tracking is available, we experiment with a synthetic setting where coarse registration is replaced by sparse ground-truth guidance, as it could be provided by perfectly accurate sparse texture tracking.
Specifically, we randomly select $N_{k}$ vertices from the ground-truth mesh, and use them to compute the distance in Equation~\ref{eq:distance} in the first stage of manifold guidance with $t > \tau$.
This replaces the coarse registration module, so coarse registration module is not used under this setting.

\begin{figure}[t]
    \includegraphics[width=0.90\linewidth]{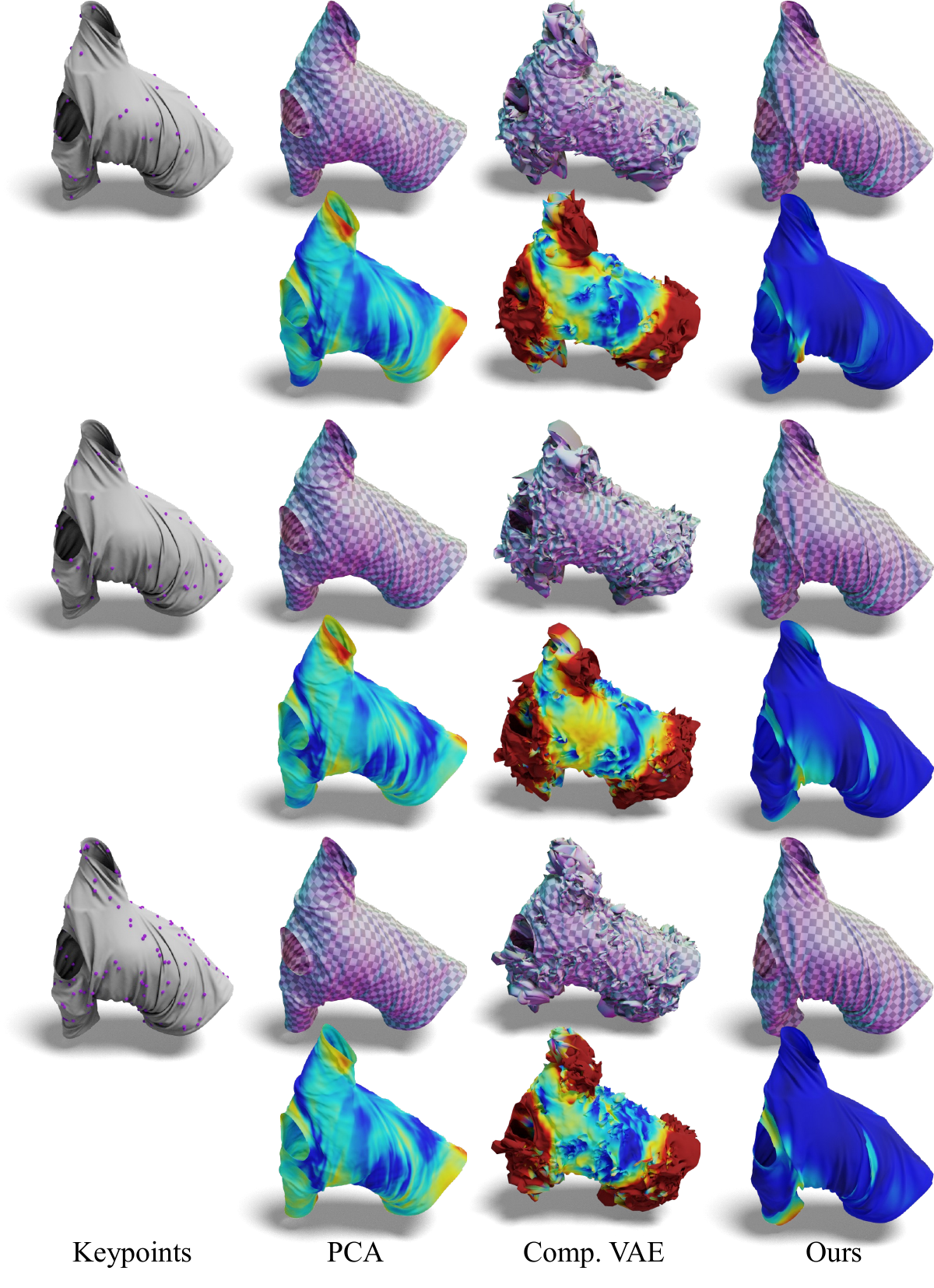}
    \centering
    \vspace{-3mm}
    \caption{Sparse ground-truth guidance. Keypoints used for tracking are shown in purple in the left column. Our method works well with very sparse tracking signal. PCA cannot reproduce the correct pose or wrinkles, and Compositional VAE fails to generate a plausible clothing shape.}
    \label{fig:application}
    \vspace{-4mm}
\end{figure}

From Table~\ref{tab:application} and Figure~\ref{fig:application}, we can see that our method performs well with very sparse keypoint tracking signals, and the accuracy improves when the number of sparse keypoints increases.
As a compact linear model, PCA performs similarly to the original setting, but increasing the number of keypoint does not help.
The compositional VAE fails to learn a meaningful latent space for plausible clothing shapes, because large deformation and fine wrinkles are coupled together.
It may require a significant amount of tracking signals to find a latent code corresponding to a plausible clothing shape.

\section{Conclusion}
\label{sec:conclusion}

We have presented a diffusion-based shape prior for highly deformable clothing geometry, and how the prior can be incorporated into fine-grained non-rigid registration tasks.
Our approach, for the first time, achieves the adoption of diffusion models into 3D cloth modeling by leveraging UV parameterization. 
Our experiments using real data show the versatility of the proposed multi-stage manifold guidance, demonstrating superior performance with multiple clothing types and diverse motions.
As our approach is simple and general, we believe it can open a new venue for various 3D optimization problems that benefit from strong 3D shape priors.

\noindent\textbf{Limitations and future work.}
As our early stage guidance relies on an off-the-shelf coarse registration module, large error introduced by this module cannot be fully removed in the following refinement stage. 
Eliminating the need of the coarse registration or building a more robust shape prior is an interesting venue for future work. 
Also, UV parameterization limits the capability of cross-garment generalization, and leveraging a single UV parameterization may be non-trivial for more complex clothing, which could be addressed by extending diffusion models to other 3D representations.

{
    \small
    \bibliographystyle{ieeenat_fullname}
    \bibliography{main}
}

\appendix
\clearpage
\setcounter{page}{1}
\maketitlesupplementary

\section{More experiment details}

\begin{figure}[t]
    \includegraphics[width=0.9\linewidth]{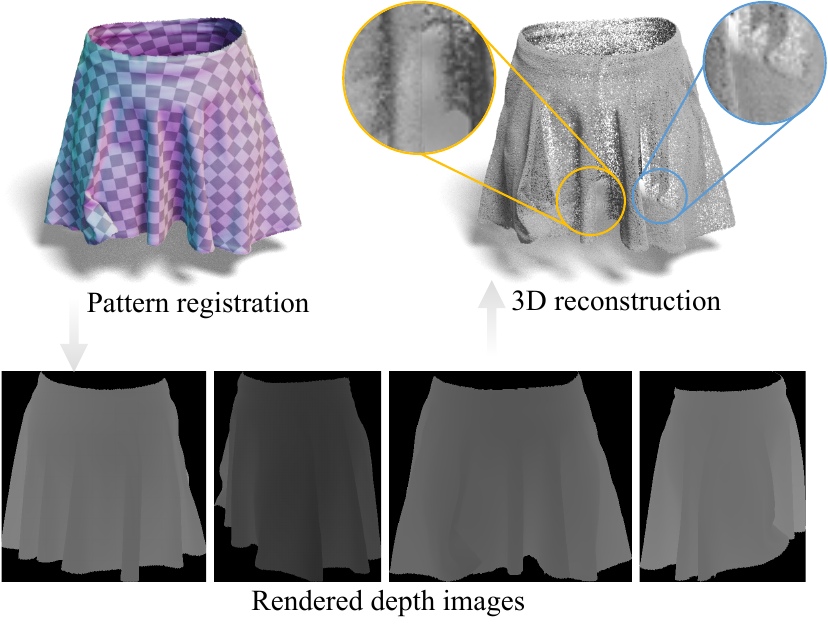}
    \centering
    \caption{We generate 3D point clouds to mimic the typical 3D reconstruction acquired by a 3D capture system. We render multi-view depth images using the ground-truth registration, then we fuse the multi-view depth images and sub-sample the 3D point cloud.}
    \label{fig:data}
    \vspace{-5mm}
\end{figure}

\noindent\textbf{Dataset.}
As illustrated in Figure~\ref{fig:data}, we render seven depth images for the ground-truth meshes from side and top views, then fuse them into a 3D point cloud with proper occlusion reasoning.

For each garment, there is a long sequence where the actor performs various movements (clips).
All clips in the sequence form the full dataset.
For extrapolation, we select a clip where the actor performs a rare movement with unique cloth deformation, not present in the rest of the sequence.
For the rest of the clips, every 20th frame is selected as interpolation set, while the remaining frames are the training set.
The number of frames in each set is shown in Table~\ref{tab:data_split}.

\begin{table}
    \centering
    \small{
    \begin{tabular}{l | c c c c } 
        \hline
        & \emph{T-shirt 1} & \emph{T-shirt 2} & \emph{Skirt 1} & \emph{Skirt 2} \\
        \cline{2-5}
        \hline
        Training & 3575 & 4463 & 1699 & 2734 \\
        Int. set & 188 & 234 & 90 & 144 \\
        Ext. set & 200 & 150 & 150 & 150 \\
        \hline
        Total & 3963 & 4847 & 1939 & 3028 \\
        \hline
    \end{tabular}
    }
    \caption{Frame count in each sequence.}
    \label{tab:data_split}
\end{table}

\noindent\textbf{Running time.}
We conduct the experiments on an NVIDIA Tesla V100 GPU with 32GB memory.
For each garment, we train the diffusion model for 100k iterations, which takes 20 hours.
The inference time is $53.57$ seconds/frame.

\noindent\textbf{SyNoRiM refinement in baseline comparison.}
We use the Adam (UniformAdam for "Lap. precond.") with step size $10^{-3}$ for 200 iterations in all the refinement experiments.
The weights for the regularization terms are $\lambda_{\text{Lap.\ reg.}} = 10^{3}$, $\lambda_{\text{Lap.\ precond.}} = 10$ (Eq.~(14) in [27]), and $\lambda_{\text{ARAP reg.}} = 10^{-3}$.

\section{Registration to noisy point cloud}
\begin{figure*}[t]
    \includegraphics[width=1\linewidth]{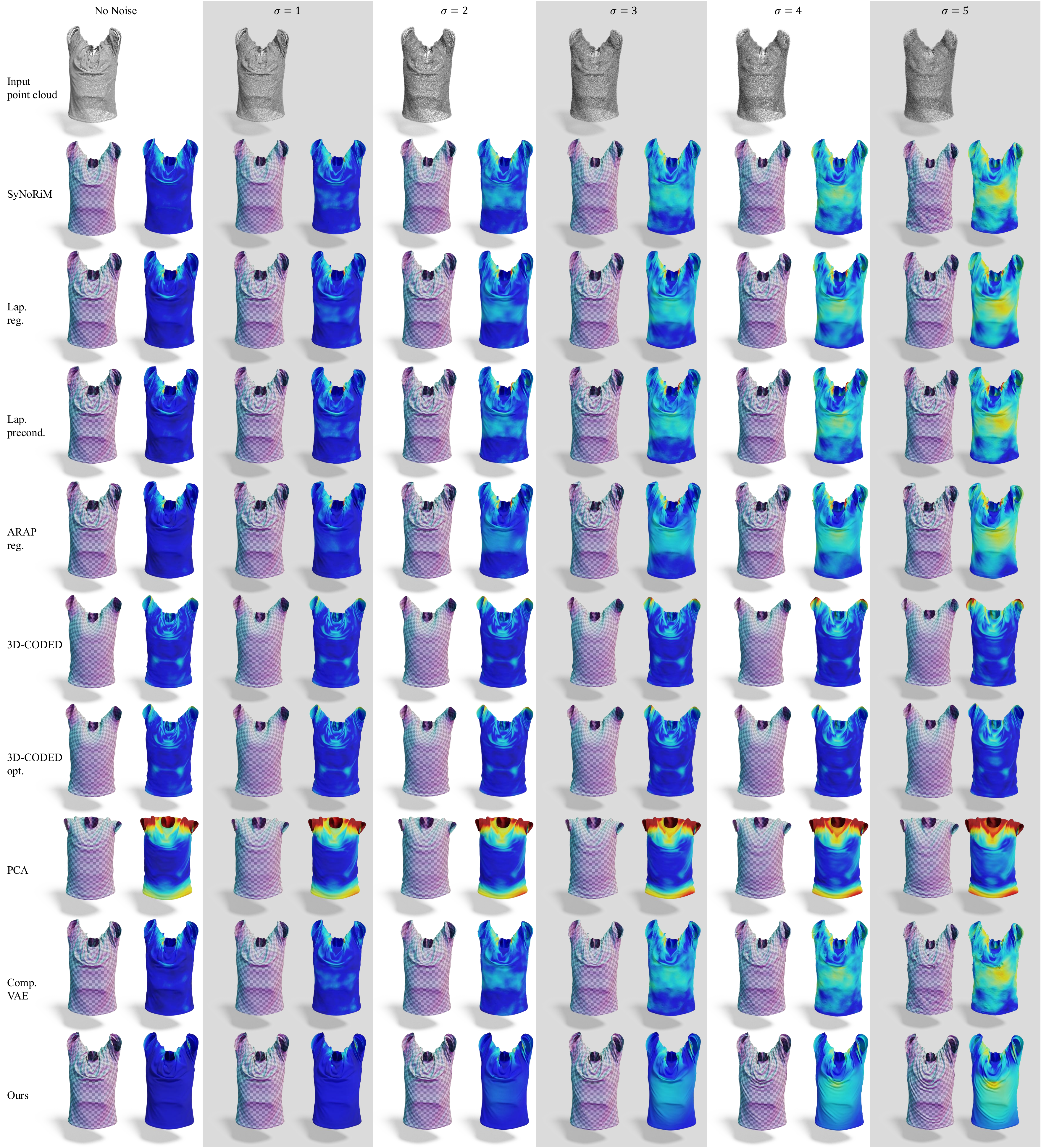}
    \centering
    \caption{Registration results of baseline methods and our method on noisy point cloud, where $\sigma$ is the standard deviation of the Gaussian noise measured in mm. Vertex error $E_{v}$ is shown in color ($0mm$ \colorbar{} $>50mm$).}
    \label{fig:noisy_comparison_gaussian}
\end{figure*}

\begin{figure*}[t]
    \includegraphics[width=0.8\linewidth]{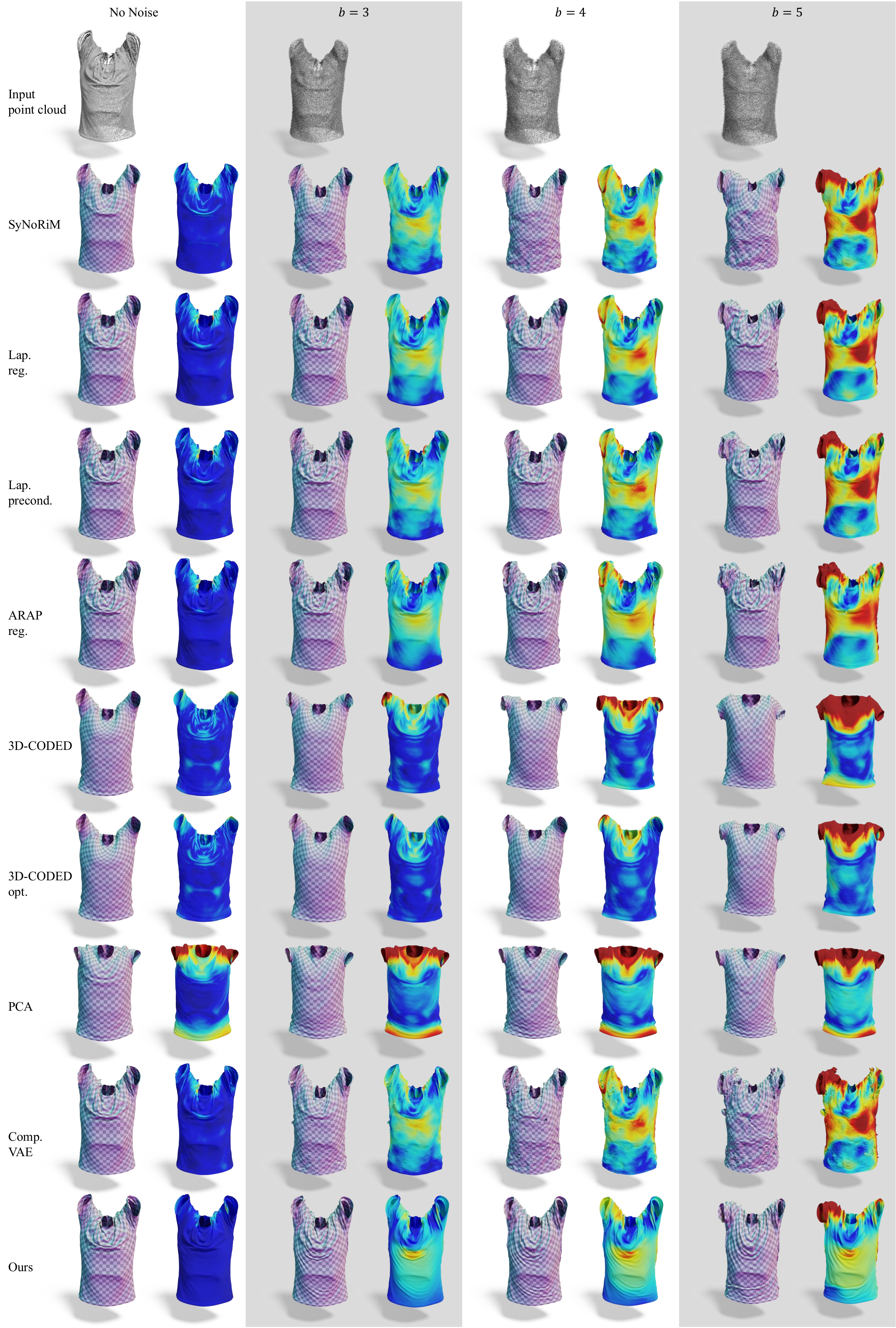}
    \centering
    \caption{Registration results of baseline methods and our method on noisy point cloud, where $b$ is the scale of the Laplace noise measured in mm. Vertex error $E_{v}$ is shown in color ($0mm$ \colorbar{} $>50mm$).}
    \label{fig:noisy_comparison_laplace}
\end{figure*}

We evaluate the robustness of our registration method to noisy measurement.
Specifically, we perturb the input point cloud of the \emph{T-shirt 1} sequence by adding Gaussian noise and Laplace noise to the 3D point locations
\begin{align}
    \mathbf{y}_{i}^{\text{G}} &= \mathbf{y}_{i} + \mathcal{N} (\mathbf{0}, \sigma^2 \mathbf{I}) \\
    \mathbf{y}_{i}^{\text{L}} &= \mathbf{y}_{i} + \mathcal{L} (\mathbf{0}, b \mathbf{I})
\end{align}
where $\mathbf{y}_{i}$ is the $i$-th point of the input point cloud $\mathcal{Y}$.
$\mathbf{y}_{i}^{\text{G}}$ is the point cloud perturbed by Gaussian noise, with $\sigma$ being the standard deviation of the Gaussian noise.
$\mathbf{y}_{i}^{\text{L}}$ is the point cloud perturbed by Laplace noise, with $b$ being the scale of the Laplace noise.
Comparing to Gaussian noise, Laplace noise has longer tail noise distribution that can mimic outliers.
In our experiments, we add Gaussian noise to the input point cloud with $\sigma = 1, 2, 3, 4, 5$ mm, and Laplace noise with $b = 3, 4, 5$ mm, as shown in the top row of Figure~\ref{fig:noisy_comparison_gaussian} and Figure~\ref{fig:noisy_comparison_laplace}.

\begin{figure*}[t]
    \includegraphics[width=0.98\linewidth]{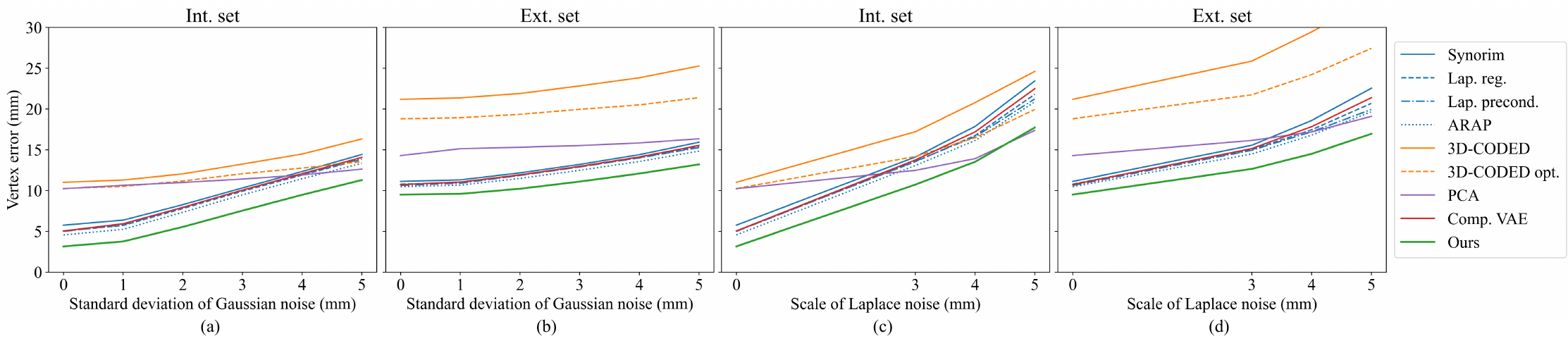}
    \centering
    \caption{Vertex error of the registration results given noisy point cloud as inputs. (a) and (b) Adding Gaussian noise to input point cloud, then testing on interpolation set and extrapolation set, respectively. (c) and (d) Adding Laplace noise to input point cloud, then testing on interpolation set and extrapolation set, respectively}
    \label{fig:noise_comparison_plot}
\end{figure*}

We take the noisy point cloud as input, and quantitatively evaluate the registration result of our method in Table~\ref{tab:noise_ours}. We also compare our method to baseline methods in Figure~\ref{fig:noise_comparison_plot}, Figure~\ref{fig:noisy_comparison_gaussian} and Figure~\ref{fig:noisy_comparison_laplace}.
The performance of our method drops with the increase of noise level, but it consistently outperforms the baseline methods.
Please note in Figure~\ref{fig:noise_comparison_plot}~(c), PCA shows comparable $E_{v}$ to our method when $b=5$. However, PCA consistently shows worse plane error $E_{pt}$ and $E_{ps}$ as discussed in the main manuscript. This can be qualitatively verified in Figure~\ref{fig:noisy_comparison_laplace} as well.

\begin{table}[t]
    \setlength{\tabcolsep}{4pt}
    \centering
    \small{
    \begin{tabular}{l | c | c  c | c  c } 
        \hline
        \multicolumn{2}{c|}{} & \multicolumn{4}{c}{\emph{T-shirt 1}} \\
        \multicolumn{2}{c|}{} & \multicolumn{2}{c|}{Int. set} & \multicolumn{2}{c}{Ext. set} \\
        \cline{3-6}
        \multicolumn{2}{c|}{} & $E_{v}$ & $E_{pt}$ / $E_{ps}$  & $E_{v}$  & $E_{pt}$ / $E_{ps}$  \\
        \hline
        \multicolumn{2}{c|}{No noise} & 3.16 & 0.57 / 0.62 & 9.51 & 0.61 / 0.75 \\
        \hline
        \multirow{5}{*}{\makecell{Gaussian \\ noise}} & $\sigma=1$ & 3.77 & 0.59 / 0.65 & 9.60 & 0.64 / 0.79 \\
        & $\sigma=2$ & 5.55 & 0.64 / 0.73 & 10.23 & 0.70 / 0.89 \\
        & $\sigma=3$ & 7.55 & 0.73 / 0.89 & 11.11 & 0.81 / 1.08 \\
        & $\sigma=4$ & 9.48 & 0.85 / 1.12 & 12.09 & 0.92 / 1.32 \\
        & $\sigma=5$ & 11.30 & 0.96 / 1.37 & 13.21 & 1.03 / 1.60 \\
        \hline
        \multirow{3}{*}{\makecell{Laplace \\ noise}} & $b=3$ & 10.74 & 0.82 / 1.10 & 12.66 & 0.89 / 1.31 \\
        & $b=4$ & 13.53 & 0.96 / 1.43 & 14.51 & 1.03 / 1.68 \\
        & $b=5$ & 17.73 & 1.11 / 1.93 & 16.96 & 1.17 / 2.14 \\
        \hline
    \end{tabular}
    }
    \caption{The quantitative evaluation of our method on noisy input point cloud. Error metrics are measured in mm.}
    \label{tab:noise_ours}
\end{table}

\section{Data prediction vs.\ noise prediction}
\begin{figure}[t]
    \includegraphics[width=\linewidth]{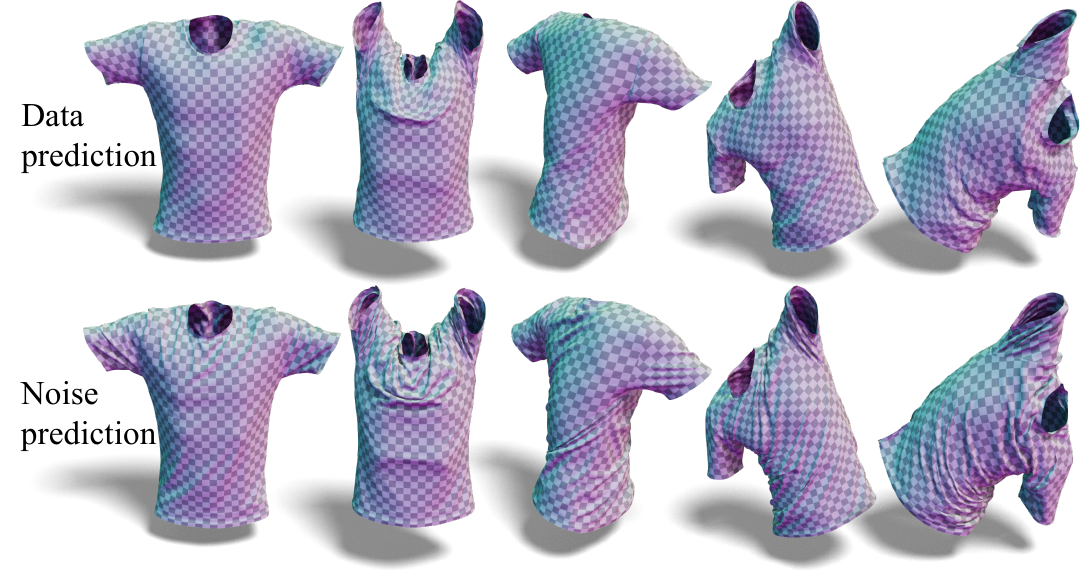}
    \centering
    \caption{The difference between predicting data $\mathbf{x}_0$ and predicting noise $\bm{\epsilon}$. Data prediction cannot represent fine details like wrinkles.}
    \label{fig:data_noise}
    \vspace{-0mm}
\end{figure}

Conceptually, it is equivalent to use a data prediction network that predicts $\mathbf{x}_0$, or a noise prediction network that predicts $\bm{\epsilon}$ in the diffusion model.
In practice, however, we find that noise prediction is more effective.
As shown in Figure~\ref{fig:data_noise}, a data prediction network has difficulty modeling high-frequency signals like wrinkles.
It cannot enforce continuity across the seams even if the seam stitching strategy is applied.

\begin{table*}[ht]
    \centering
    \small{
    \begin{tabular}{l | c  c  | c  c  | c  c  | c  c  } 
        \hline
        Validation subject & \multicolumn{2}{c|}{subject\_00} & \multicolumn{2}{c|}{subject\_01} & \multicolumn{2}{c|}{subject\_02} & \multicolumn{2}{c}{subject\_03} \\
        \cline{2-9}
        & $E_{v}$ & $E_{pt}$/$E_{ps}$  & $E_{v}$  & $E_{pt}$/$E_{ps}$  & $E_{v}$  & $E_{pt}$/$E_{ps}$  & $E_{v}$  & $E_{pt}$/$E_{ps}$  \\
        \hline
        SyNoRiM~\cite{huang2022multiway} & 10.62 & 1.38 / 1.66 & 13.59 & 1.54 / 2.16 &  8.50 &  1.33 / 1.57 & 10.67 & 1.40 / 1.80 \\
        Lap. reg.~\cite{sorkine2004laplacian} & 10.32 & 0.66 / 0.64 & 12.99 & 0.64 / 0.67 & 8.13 & 0.61 / 0.62 & 10.21 & 0.65 / 0.65 \\
        Lap. precond.~\cite{nicolet2021large} & 10.09 & 0.53 / 0.59 & 12.97 & 0.55 / 0.63 & 8.10 & 0.52 / 0.58 & 10.20 & 0.56 / 0.62 \\
        ARAP reg.~\cite{sorkine2007rigid} & 10.04 & 0.52 / 0.60 & 12.57 & 0.54 / 0.65 & 7.87 & 0.51 / 0.58 & 9.94 &  0.54 / 0.65 \\
        \hline
        3D-CODED~\cite{groueix20183d} & 36.94 & 3.93 / 15.77 & 18.17 & 4.09 / 6.05 & 23.91 & 4.03 / 6.76 & 15.80 & 3.25 / 4.58 \\        
        3D-CODED opt. & 33.03 & 3.13 / 13.13 & 16.04 & 3.00 / 4.17 & 20.52 & 3.35 / 5.34 & 14.14 & 2.77 / 3.72 \\
        \hline
        PCA & 11.65 & 3.02 / 3.44 & 12.11 & 2.72 / 3.28 & 10.19 & 2.87 / 3.27 & 10.41 & 2.38 / 2.71 \\
        \hline
        Comp. VAE~\cite{bagautdinov2018modeling} & 10.35 & 1.21 / 1.23 & 13.12 & 1.06 / 1.19 & 8.21 & 1.05 / 1.08 & 10.27 & 0.97 / 1.04 \\
        \hline
        Ours & \textbf{9.85} & 0.61 / 0.69 & \textbf{11.77} & 0.60 / 0.75 & \textbf{7.58} & 0.56 / 0.68 & \textbf{9.69} & 0.60 / 0.72 \\
        \hline
    \end{tabular}
    }
    \vspace{2mm}
    \caption{Quantitative results for cross-subject generalization. For each validation, the models are trained on the other 3 subjects. Error metrics are measured in mm. \textbf{Bold} indicates the best $E_{v}$.}
    \label{tab:cross-subject-4-fold}
\end{table*}

\section{Cross-subject generalization}
\begin{figure*}[t]
    \includegraphics[width=0.96\linewidth]{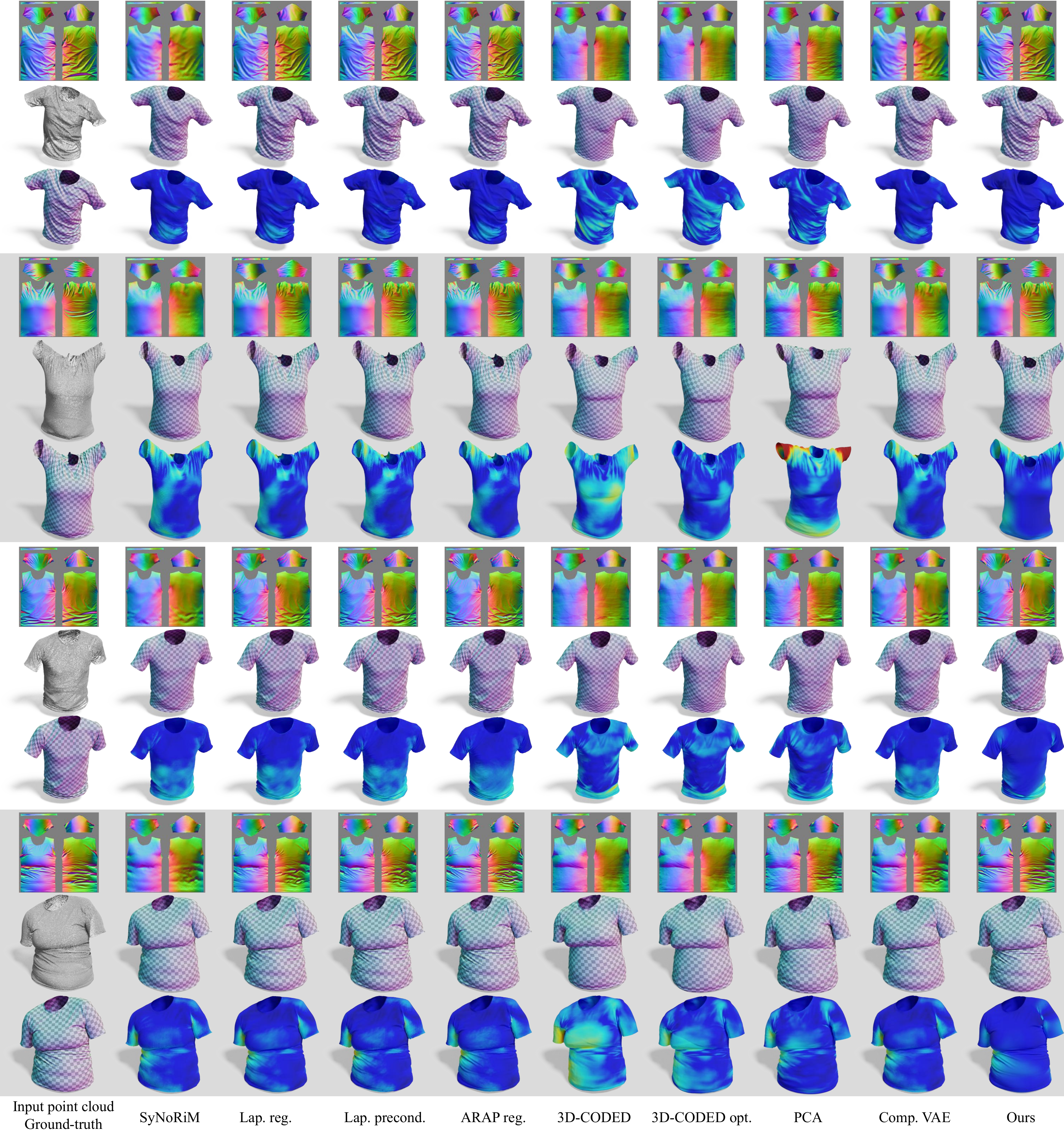}
    \centering
    \caption{Qualitative comparison for cross-subject generalization. Each example is from a different subject. For each example, the model is trained on the other 3 subject, so the test subject is always unseen during training. See Figure~\ref{fig:comparison_tshirt_1} for explanation of the figure.}
    \label{fig:comparison-cross_subject}
    \vspace{-3mm}
\end{figure*}

The same clothing worn by different subjects may deform differently because of the variation of body shapes.
In Table~\ref{tab:cross-subject-4-fold} and Figure~\ref{fig:comparison-cross_subject}, we empirically show that the proposed method can generalize to unseen subjects.
Specifically, \emph{T-shirt 1} is worn by 4 different actors in 4 long sequences "subject\_00", "subject\_01", "subject\_02", and "subject\_03".
In this experiment, we take \emph{T-shirt 1} on all subjects as the full dataset, and divide it into 4 parts, each contains one subject.
We conduct 4-fold cross-validation, where each time we use 3 subjects for training, and the 4th subject for validation.
In each validation, the training set consists of all frames of the 3 subjects, while the validation set only contains every 20th frame of the 4th subject.
In Table~\ref{tab:cross-subject-4-fold}, we report the quantitative evaluation of the 4-fold cross-validation, showing that the proposed method consistently outperforms the baseline methods on the cross-subject generalization task.

Please note that \emph{T-shirt 1} on "subject\_01", "subject\_02", and "subject\_03" are only used in the cross-subject generalization experiment, while other experiments on \emph{T-shirt 1} are done with "subject\_00".

\section{Additional qualitative results}
\begin{figure*}[t]
    \includegraphics[width=0.85\linewidth]{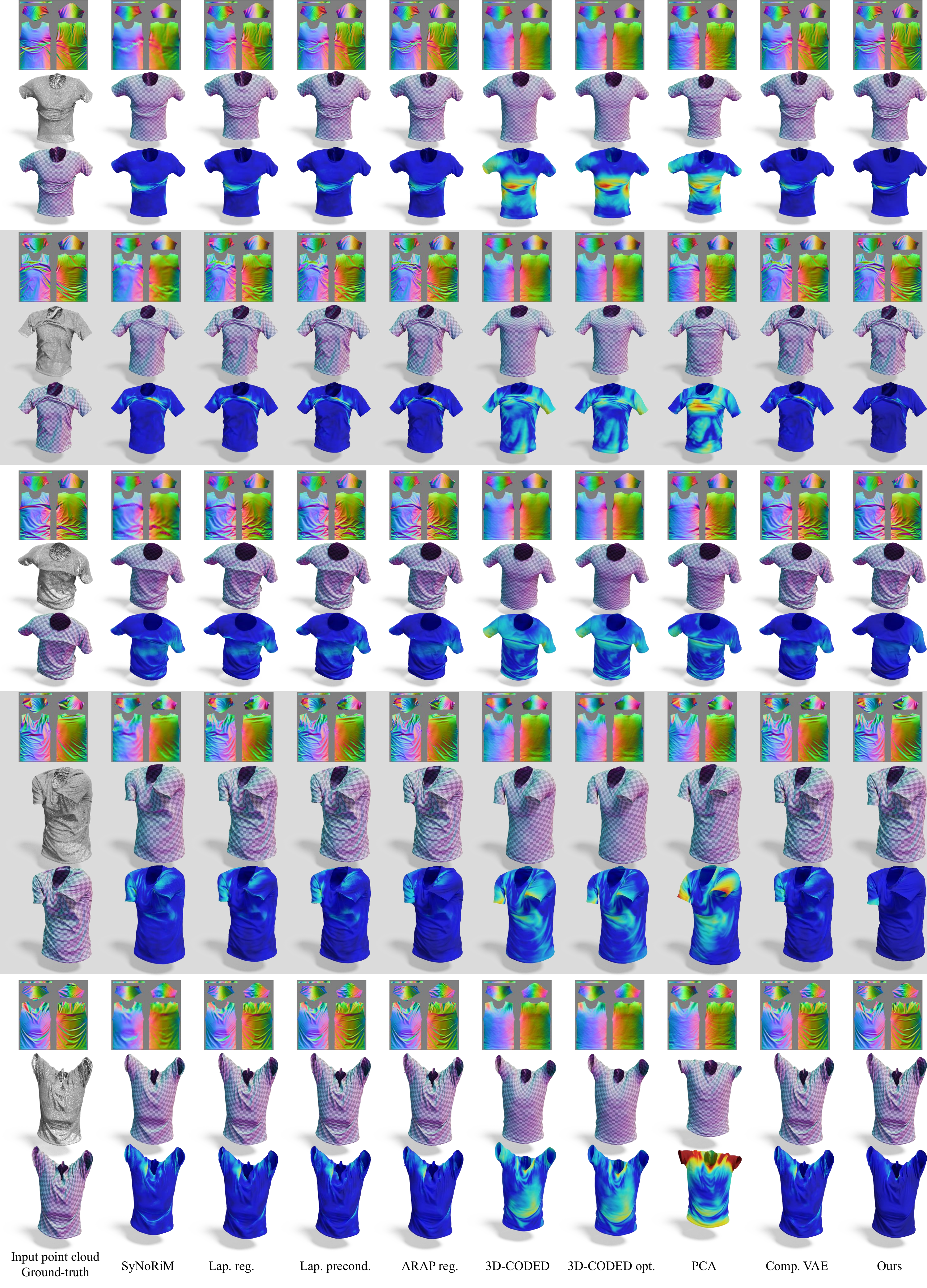}
    \centering
    \caption{Comparison to baseline methods on \emph{T-shirt 1}. In each example, the middle-left is the input point cloud, the bottom-left is the ground-truth, the top-left is normal map of ground-truth. The rest are the results of different methods, where the top row shows normal map, the middle row shows the geometry with normal rendering, while the bottom row shows vertex error $E_{v}$ in color ($0mm$ \colorbar{} $>50mm$).}
    \label{fig:comparison_tshirt_1}
\end{figure*}

\begin{figure*}[t]
    \includegraphics[width=0.85\linewidth]{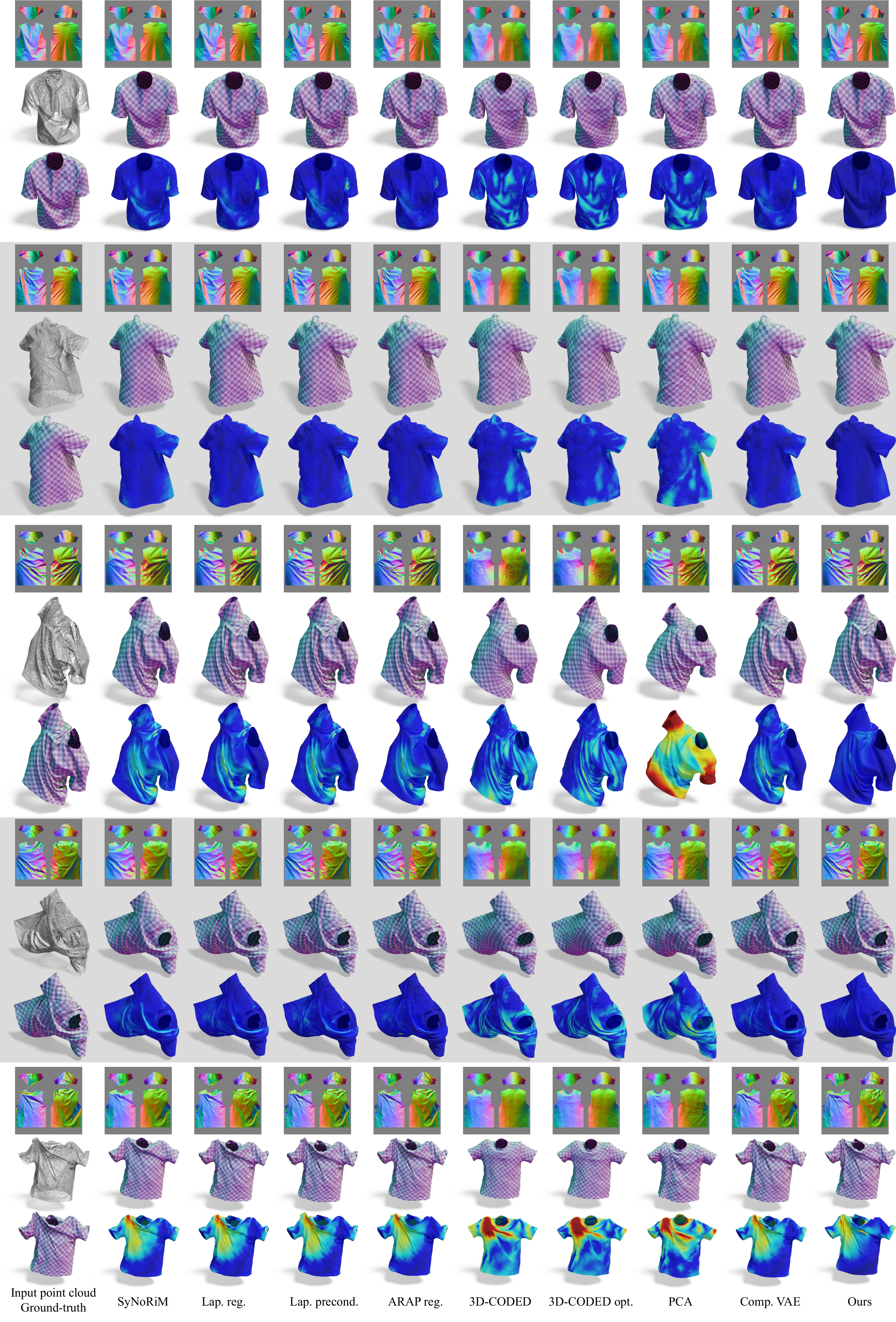}
    \centering
    \caption{Comparison to baseline methods on \emph{T-shirt 2}. See Figure~\ref{fig:comparison_tshirt_1} for explanation.}
    \label{fig:comparison_tshirt_2}
\end{figure*}

\begin{figure*}[t]
    \includegraphics[width=0.9\linewidth]{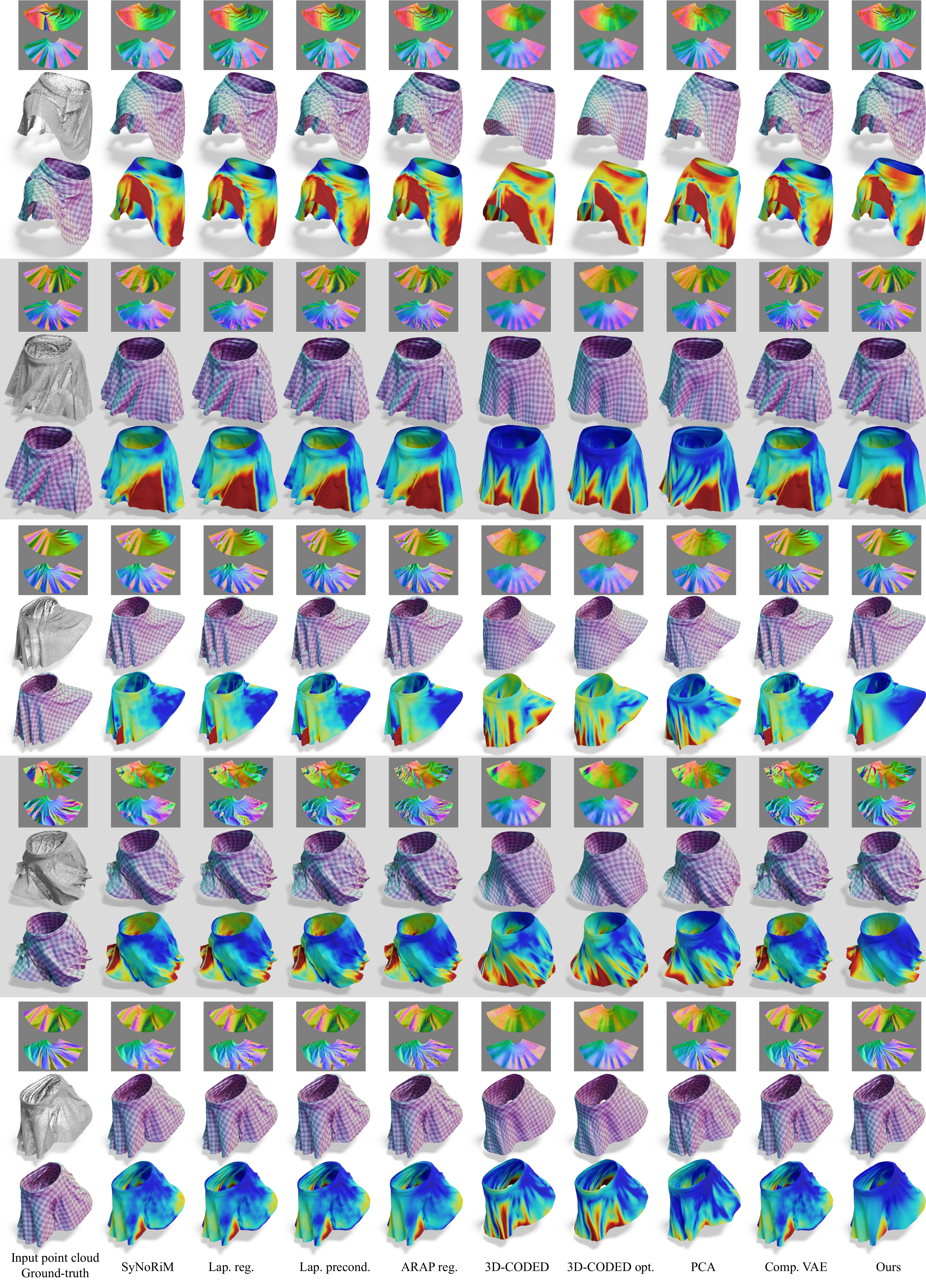}
    \centering
    \caption{Comparison to baseline methods on \emph{Skirt 1}. See Figure~\ref{fig:comparison_tshirt_1} for explanation.}
    \label{fig:comparison_skirt_1}
\end{figure*}

\begin{figure*}[t]
    \includegraphics[width=0.8\linewidth]{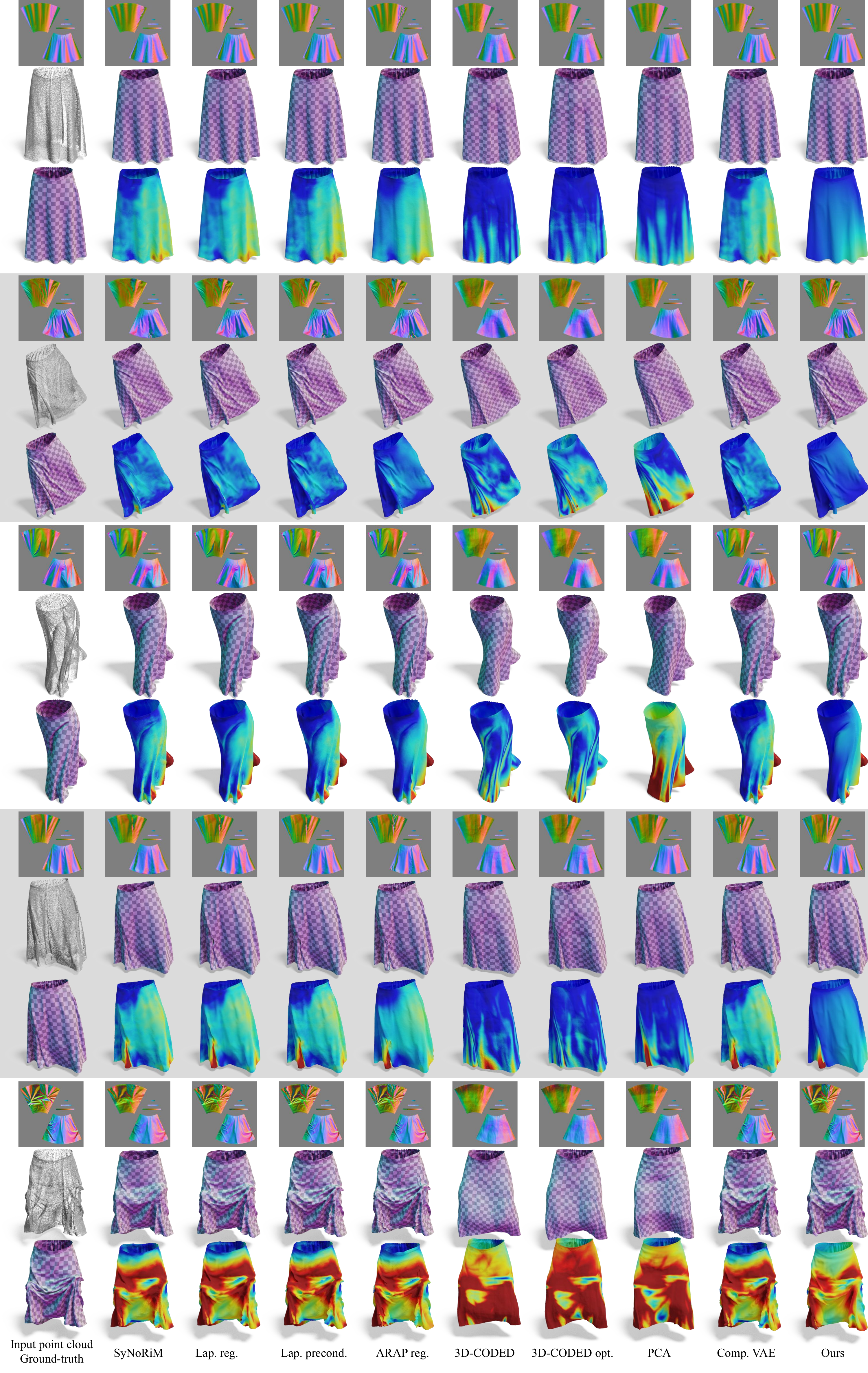}
    \centering
    \caption{Comparison to baseline methods on \emph{Skirt 2}. See Figure~\ref{fig:comparison_tshirt_1} for explanation.}
    \label{fig:comparison_skirt_2}
\end{figure*}

We show more qualitative results on \emph{T-shirt 1}, \emph{T-shirt 2}, \emph{Skirt 1} and \emph{Skirt 2} sequences in Figure~\ref{fig:comparison_tshirt_1}, Figure~\ref{fig:comparison_tshirt_2}, Figure~\ref{fig:comparison_skirt_1} and Figure~\ref{fig:comparison_skirt_2}, respectively.

\end{document}